%% file: main.tex
\documentclass[10pt,twocolumn,letterpaper]{article}

\usepackage[pagenumbers]{cvpr} 

\input{preamble}

\input{math_commands.tex}
\usepackage{adjustbox}
\usepackage[shortlabels]{enumitem}

\usepackage{algorithm}
\usepackage{algpseudocode}

\usepackage{newfloat}
\usepackage{listings}

\definecolor{cvprblue}{rgb}{0.21,0.49,0.74}
\usepackage[pagebackref,breaklinks,colorlinks,allcolors=cvprblue]{hyperref}

\def\paperID{7} 
\def\confName{CVPR}
\def\confYear{2026}

\title{\vspace{-1.0cm}\ours{}: Part-Aware Multi-View Diffusion \\ for Creative Fine-Grained Object Generation \vspace{-0.4cm}}

\author{
Kam Woh Ng$^{1}$ \quad
Jing Yang$^{2}$ \quad
Jia Wei Sii$^{3}$ \quad
Chee Seng Chan$^{3}$ \quad
Jiankang Deng$^{4}$ \\
Yi-Zhe Song$^{1}$ \quad
Tao Xiang$^{1}$ \quad
Xiatian Zhu$^{1}$ \\
\\
$^{1}$University of Surrey \quad
$^{2}$University of Cambridge \quad
$^{3}$Universiti Malaya \quad
$^{4}$Imperial College London \\
\url{https://github.com/kamwoh/chirpy3d}
}

\begin{document}

\input{secs/0_abstract}

\input{secs/0_intro}
\input{secs/1_related}

\input{secs/2_method}
\input{secs/3_exp}

\input{secs/4_conclusion}

{
    \small
    \bibliography{reference}
    \bibliographystyle{ieeenat_fullname}
}

\input{supp}

\end{document}

%% file: preamble.tex
\usepackage[dvipsnames]{xcolor}

\newcommand{\ours}{Chirpy3D} 
\newcommand*{\method}{\ours{}}
\newcommand{\PC}{PartCraft} 
\newcommand{\TI}{Textual Inversion}
\usepackage{amssymb}
\usepackage{pifont}
\newcommand{\cmark}{\ding{51}}
\newcommand{\xmark}{\ding{55}}

%% file: math_commands.tex
\usepackage{amsmath,amsfonts,bm}

\def\eqref#1{equation~\ref{#1}}

\def\1{\bm{1}}

\DeclareMathAlphabet{\mathsfit}{\encodingdefault}{\sfdefault}{m}{sl}
\SetMathAlphabet{\mathsfit}{bold}{\encodingdefault}{\sfdefault}{bx}{n}

\usepackage{mathtools}

%% file: secs/0_abstract.tex
\twocolumn[{
\renewcommand\twocolumn[1][]{#1}
\maketitle
\begin{center}
    \centering
    \vspace{-0.7cm}
    \includegraphics[width=\textwidth]{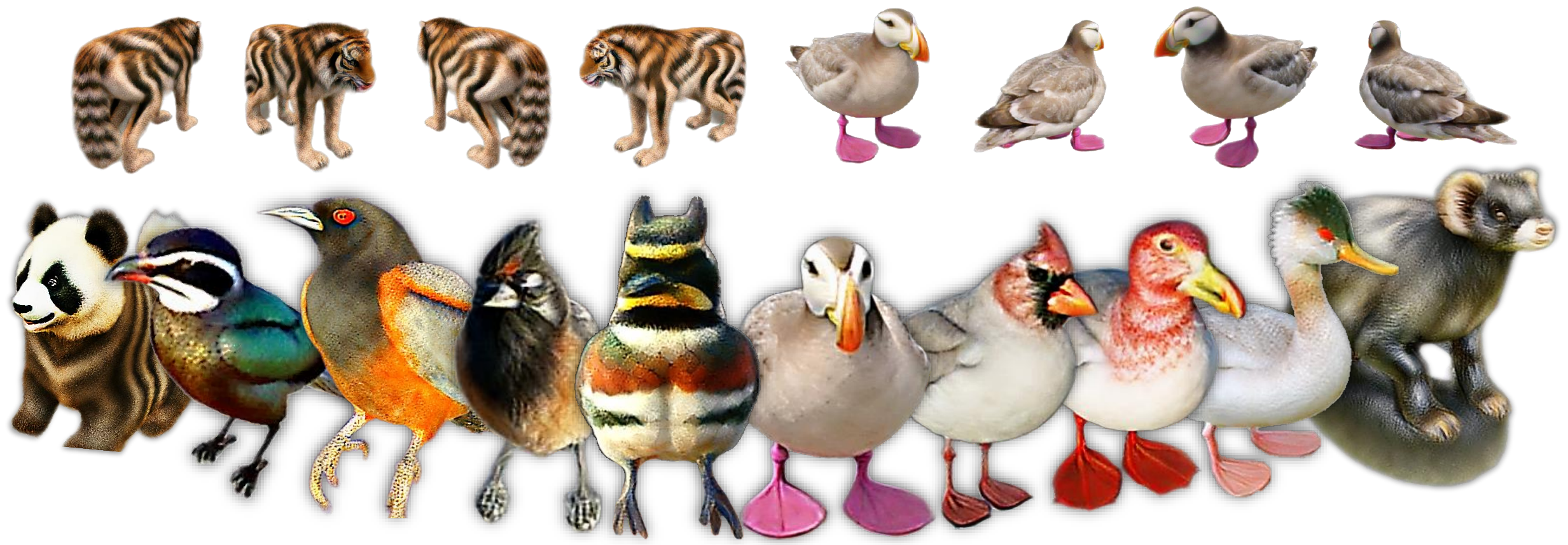}
    \captionof{figure}{
    \textbf{Creative 3D creatures generated by \textit{Chirpy3D}.} The creativity is achieved through part-level sampling in a learned latent space. The selected parts are composed into a textual prompt, which is then fed into a multi-view diffusion model to guide the 3D representation optimization (NeRF \cite{mildenhall2020nerf}) via SDS loss \cite{poole2022dreamfusion}.
    }
    \label{fig:big-teaser}
\end{center}
}]

\begin{abstract}

Understanding and generating the fine-grained structure of objects -- such as birds with species-specific beaks, wings, and tails -- is a long-standing challenge in computer vision.
We propose \ours{}, a part-aware multi-view diffusion framework that learns a hierarchical part latent space from unposed 2D images, using only off-the-shelf 2D part segmentation masks as spatial guidance -- without requiring any 3D data, camera poses, or manual part annotations. This latent space enables intuitive part-level swapping, interpolation, and zero-shot composition. A self-supervised feature consistency loss further encourages structural alignment across views, allowing coherent generation even with hybrid or unseen part combinations.
Our core contribution is the controllable part-aware latent space and multi-view diffusion model. Downstream 3D generation is supported via any differentiable renderer such as NeRF but is orthogonal to the main framework, making Chirpy3D a flexible foundation for creative object generation in the absence of structured 3D data.

\end{abstract}

%% file: secs/0_intro.tex
\section{Introduction}\label{sec:intro}

Understanding and modeling the fine-grained structure of objects is a long-standing challenge in computer vision and graphics~\cite{bourdev2009poselets, felzenszwalb2008discriminatively, krause2015fine, hinton2023represent, zhao2021part, peng2017object, ng2024concepthash, ng2024partcraft, han2024headsculpt}. Many real-world objects -- such as animals~\cite{he2021partimagenet}, vehicles \cite{krause20133d}, and furniture \cite{chang2015shapenet} -- share global anatomical structures while exhibiting rich part-level variation. For example, birds~\cite{wah2011caltech} may differ in beak shapes, wing patterns, or tail geometries depending on species. Learning such part-aware representations opens up exciting opportunities for controllable image and 3D object generation (see Figure~\ref{fig:big-teaser}).

Diffusion models have recently driven major progress in both 2D~\cite{rombach2022high, saharia2022photorealistic, du2024demofusion} and 3D generation~\cite{gao2022get3d, wu2024direct3d, xiang2025structured}. In particular, multi-view diffusion models~\cite{liu2023zero, shi2023zero123plus, liu2023one2345, liu2024one, lai2025hunyuan3d, yang2024hunyuan3d, kong2024eschernet, kant2024spad, shi2023mvdream} have demonstrated the ability to synthesize 3D-consistent images from a single view. More recent approaches such as PartGen~\cite{chen2025partgen} go further by introducing explicit part-level supervision for multi-view 3D generation via segmentation and completion pipelines. However, most existing methods are trained on curated 3D datasets~\cite{deitke2023objaverse, wu2023omniobject3d} and focus on general-purpose generation. They lack the flexibility for part-wise editing, creative recombination, or structural control through latent space.

In contrast, learning part-aware representations for fine-grained categories (such as novel birds with hybrid, species-specific traits) remains underexplored, particularly in settings where 3D annotations or posed multi-view images are unavailable. PartCraft~\cite{ng2024partcraft} pioneered 2D part-aware generation by learning discrete per-class textual embeddings for each part, but these do not form a continuous latent space -- preventing smooth interpolation, random sampling, or generalization to unseen combinations, and it operates only in 2D. We extend this by learning a \textit{continuous} part latent space with a Gaussian prior, enabling structured interpolation and sampling, integrated with a multi-view diffusion backbone for downstream 3D generation.

In this paper, we focus on learning a \textbf{part-aware, multi-view diffusion model} trained on \textbf{unposed, unlabeled 2D images} from fine-grained object categories, without 3D ground truth, camera poses, or keypoint annotations. To guide part-attention learning, we use 2D part segmentation masks from an off-the-shelf detector~\cite{ng2024partcraft} as spatial guidance. Our key idea is to learn a \textbf{hierarchical part latent space} enabling compositional control over object parts, supporting operations such as interpolation, swapping, and zero-shot composition. We further introduce a \textbf{self-supervised feature consistency loss} that encourages structural alignment across views, helping generate coherent shapes even with hybrid or unseen parts. While we demonstrate 3D generation via NeRF~\cite{mildenhall2020nerf} and 3DGS~\cite{kerbl20233dgs}, the choice of 3D backend is orthogonal -- our core contribution is the controllable part-aware latent space.

\input{pdfs/fig2-overview}

%% file: pdfs/fig2-overview.tex
\begin{figure*}[t]
    \centering
    \includegraphics[width=\linewidth]{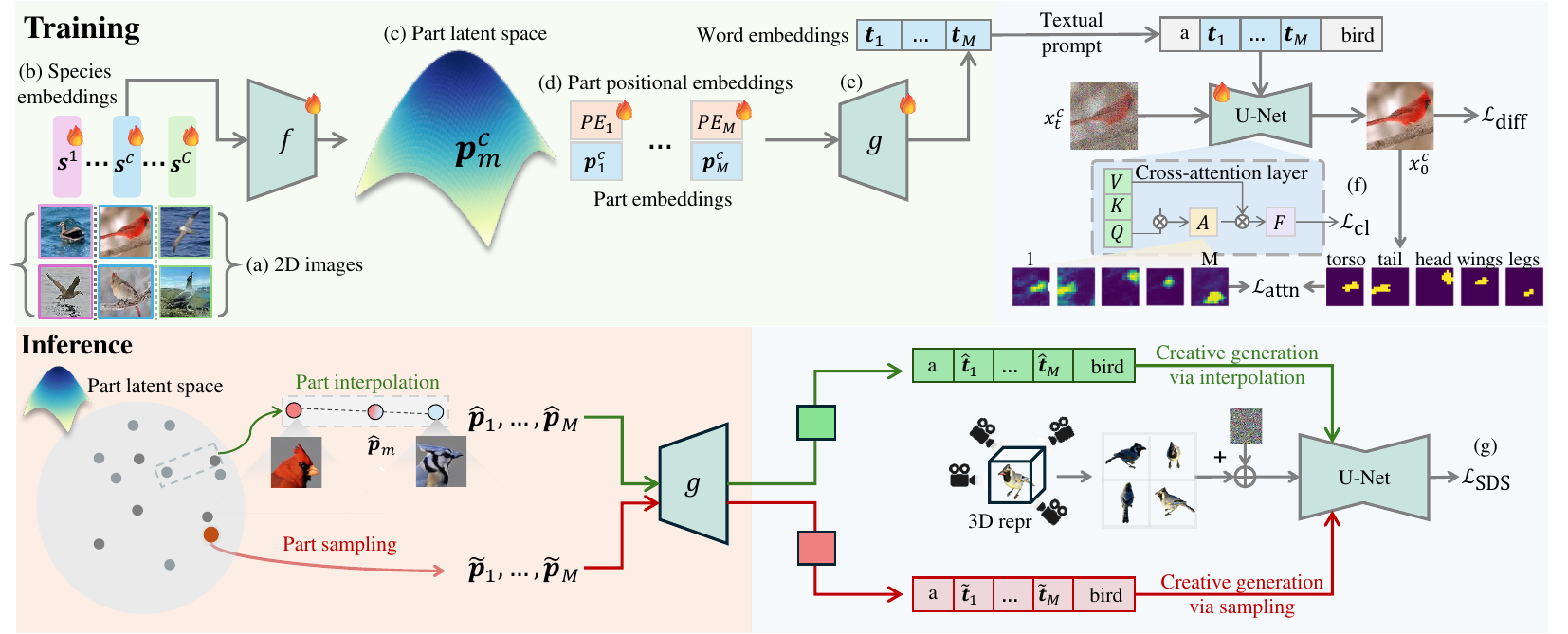}
    \caption{
    Overview of \textbf{\ours{}}.  
\ours{} takes \textbf{(a)} a set of unposed 2D images from multiple fine-grained species (e.g., birds) and \textbf{(b)} learns to decompose each object into a set of underlying parts (e.g., head, wings, torso, legs, tail) within a hierarchical part latent space -- species embedding $\mathbf{s}$ captures global species characteristics while part-level embedding $\mathbf{p}$ captures fine-grained part variations.
\textbf{(c-d)} The part latent space is regularized with a Gaussian prior for smooth interpolation and novel part sampling, with shared part-specific position embeddings ($PE$) enabling cross-species alignment.
\textbf{(e)} Part embeddings are projected via a learnable function $g$ into part-aware textual embeddings to condition the multi-view diffusion model (\eg, MVDream \cite{shi2023mvdream}) to generate multi-view images.
\textbf{(f)} A self-supervised feature consistency loss is applied to enforce structural and semantic coherence across views, improving the realism and alignment of both seen and unseen parts.  
\textbf{(g)} During inference, \ours{} supports both reconstruction and creative generation -- either directly using learned part latent codes or sampling/interpolating within the part latent space -- which guides 3D representation learning (\eg, NeRF or 3DGS) via SDS optimization. 
    }
    \label{fig:overview}
\end{figure*}

%% file: secs/1_related.tex
\section{Related work}

\noindent \textbf{Fine-grained generation.}  
Research on fine-grained visual understanding has primarily focused on 2D tasks, including fine-grained recognition \cite{zheng2019looking,chang2020devil,du2020fine,wei2021fgsurvey,ng2024concepthash} and 2D object generation \cite{singh2019finegan, xu2018attngan, li2021partgan, ding2022comgan, li2020mixnmatch, chen2022sphericgan}. In contrast, fine-grained 3D generation remains underexplored, with most existing works focusing on relatively simple, rigid categories such as chairs and cars \cite{yang2024sym3d,gao2022get3d,siddiqui2024meshgpt,hui2022neural,genova2020local,genova2019learning,park2019deepsdf}. Deformable fine-grained categories, such as birds and dogs, pose unique challenges due to complex textures, diverse part configurations, and high pose variability, while obtaining 3D data remains expensive. Our work pushes beyond these limitations to enable the creative synthesis of novel, fine-grained objects directly from unposed 2D images.

\noindent \textbf{Diffusion models.}  
Diffusion models have recently been applied to multi-view generation, though they often suffer from multi-view inconsistencies — commonly known as the ``Janus problem.'' Text-to-multi-view methods such as MVDream \cite{shi2023mvdream} and SPaD \cite{kant2024spad} generate object views from text prompts, while methods like Zero123 \cite{liu2023zero}, Zero123++ \cite{shi2023zero123plus}, and EscherNet \cite{kong2024eschernet} infer novel views from a single image. These methods focus on rendering consistent views of known objects but lack the ability to model fine-grained part-level variations or creative generation of novel objects. Our approach extends this line of work by enabling the generation of entirely new fine-grained objects in a zero-shot setting, including novel part combinations and species-level diversity.

\noindent \textbf{Part-aware object generation.}  
Part-aware generation has been extensively explored in 2D, where methods \cite{avrahami2023breakascene, ng2024partcraft} such as PartCraft~\cite{ng2024partcraft} leverage attention-based objectives to disentangle and recombine object parts. However, these methods typically operate on predefined or salient parts of specific object categories and do not model a continuous part distribution or enable generalizable part composition across diverse instances. In 3D, most part-aware approaches rely on synthetic datasets with explicit part annotations (e.g., ShapeNet~\cite{chang2015shapenet}), limiting their applicability to fine-grained or deformable categories~\cite{wu2019sagnet, hertz2022spaghetti, koo2023salad, park2019deepsdf, jang2021codenerf}. More recently, PartGen~\cite{chen2025partgen} can decompose generated shapes into parts post hoc using supervised segmentation and completion pipelines but does not model part representations explicitly during generation.
We break these limitations by enabling the creative generation of deformable fine-grained objects without 3D supervision, camera poses, or manual part labels -- using only 2D part segmentation masks from an off-the-shelf detector~\cite{ng2024partcraft} as weak spatial guidance for attention learning, while modeling parts as continuous distributions for flexible recombination. Unlike prior creative synthesis methods based on GANs~\cite{ahmed2017can, nobari2021creativegan, ge2020doodlergan}, VAEs~\cite{das2020creativedecoder, cintas2022creativity}, or diffusion-based concept mixing~\cite{vinker2023conceptdecomposition, richardson2023conceptlab, li2024tp2o}, which rely on discrete token or embedding composition, our approach learns a continuous part latent space that supports structured sampling, interpolation, and compositional generation.

%% file: secs/2_method.tex
\section{Methodology}

\subsection{Overview}

Our goal is to learn a part-aware generative model from unposed 2D images of fine-grained categories (e.g., bird species), without 3D ground truths\footnote{Rendered views from 3D meshes.} or camera poses, enabling creative object synthesis and downstream 3D generation in a zero-shot setting. To achieve this, we propose \textbf{\ours{}} (see Figure~\ref{fig:overview}), a multi-view diffusion framework that learns a \textit{hierarchical part latent space}, where both species-level and part-level information are represented in a structured and compositional manner. Part embeddings are projected into a space compatible with a pre-trained multi-view diffusion model (e.g., MVDream~\cite{shi2023mvdream}), enabling the model to both reconstruct seen species and compose novel objects with unseen part combinations.

\filbreak
\subsection{Hierarchical part latent space}\label{sec:part_latent_space}

\ours{} represents each object as a set of $M$ semantic parts (e.g., $M = 5$ for head, wings, torso, legs, and tail). We follow the part definition used by the off-the-shelf segmentor~\cite{ng2024partcraft}, which provides consistent spatial masks for these five regions; exploring adaptive or variable part counts is left for future work. For a dataset spanning $C$ species, we learn a species-level embedding for each class, denoted by $\{\mathbf{s}^c \in \mathbb{R}^{D_s}\}_{c=1}^C$. These embeddings serve as compact encodings of species identity and are mapped to part embeddings via a learnable projection function $f$:
\begin{align}
    \{\mathbf{p}^c_m \in \mathbb{R}^{D_p}\}_{m=1}^M = f(\mathbf{s}^c).
\end{align}
To encourage a smoother interpolation between species and enable sampling of novel parts, we regularize the part embeddings with a standard Gaussian prior:
\begin{align} \label{eq:reg}
    \mathcal{L}_{\text{reg}} = \mathbb{E}_{c,m} \left[ \| \mathbf{p}^c_m \|^2 \right].
\end{align}
Since all part embeddings are decoded by a shared projection function $g$ into the same semantic space (e.g., a CLIP-style textual embedding space), we introduce learnable part-specific positional encodings $\{ \text{PE}_m \in \mathbb{R}^{D_p} \}_{m=1}^M$ to distinguish between part types. Each part embedding is concatenated with its corresponding positional encoding before decoding:
\begin{align}
    \mathbf{t}^c_m = g\left( \left[ \mathbf{p}^c_m, \text{PE}_m \right] \right), \quad \forall m = 1, \dots, M,
\end{align}
where $[\cdot, \cdot]$ denotes vector concatenation. Functions $f$ and $g$ are implemented as a one-layer and two-layer MLP, respectively.
This design maintains a unified decoding space while preserving semantic distinctions between parts, enabling part mixing and recombination across species.

\subsection{Model optimization}
We fine-tune a multi-view diffusion backbone using the standard diffusion loss \cite{ho2020ddpm}:
\begin{align} \label{eq:MV}
    \mathcal{L}_{\text{diff}} &= \mathbb{E}_{\bm{x}, \bm{y}, t, \epsilon}\left[ \Vert \epsilon - \epsilon_\theta(\bm{z}_t; \bm{y}, t) \Vert_{2}^{2} \right],
\end{align}
with $\bm{z}_t$ representing the noised latent state of input $\bm{x}$ at timestep $t$ and $\bm{y}$ as the embeddings\footnote{As standard CLIP textual embeddings, the shape is $77 \times D_t$.} of text prompt like ``a $\bm{t}^c_{1}$,...,$\bm{t}^c_{M}$ bird".
To enforce part-wise disentanglement during generation, we adopt the entropy-based attention loss from~\cite{ng2024partcraft}. This loss encourages each part textual embedding $\bm{t}^c_m$ to attend to its corresponding spatial location in the cross-attention maps during fine-tuning of the diffusion backbone:
\begin{align}
    \mathcal{L}_{\text{attn}} &= \mathbb{E}_{\bm{z},t,m} \big[ -\mathbb{S}_{m} \log \hat{\mathbb{A}}_m \big], \label{eq:attn} \\
    \quad \textit{where} \quad \hat{\mathbb{A}}_{m,i,j} &= \frac{\bar{\mathbb{A}}_{m,i,j}}{\sum_{k=1}^{M} \bar{\mathbb{A}}_{k,i,j}}, \quad  \bar{\mathbb{A}}_m = \frac{1}{L}\sum_l^L \mathbb{A}_{l,m}, \nonumber
\end{align}
where $\mathcal{L}_{\text{attn}}$ is the cross-entropy loss between the normalized cross-attention map of part $m$ and its binary segmentation mask $\mathbb{S}_m$, obtained from the part segmentation module in \cite{ng2024partcraft}.
Here, $\mathbb{A}_{l,m}$ represents the cross-attention map from layer $l$, indicating the correlation between part $m$ and the noisy latent $\bm{z}_t$. The averaged attention map $\bar{\mathbb{A}}_m$ aggregates attention across $L$ selected layers, while $\hat{\mathbb{A}}_{m,i,j}$ normalizes the attention across all $M$ parts. Note that our framework requires no 3D ground truth or camera poses. The only external input is 2D part segmentation masks from an off-the-shelf detector~\cite{ng2024partcraft}, which provide spatial guidance for attention learning. Importantly, these masks are not treated as ground-truth part labels but rather as a weak supervisory signal that encourages each part embedding to attend to its corresponding spatial region.

\input{pdfs/fig3-regloss}

\noindent
\textbf{Self-supervised feature consistency loss.}
To enhance structural and semantic coherence across views, we introduce a self-supervised feature consistency loss $\mathcal{L}_{cl}$, which improves the visual quality and geometric consistency of both seen and unseen (novel) objects.
The key intuition is that for a given part composition, the cross-attention features -- which determine \textit{what} content is generated and \textit{where} -- should remain stable regardless of the specific noise realization. If different noise inputs at the same timestep produce divergent internal features, the model is sensitive to stochastic noise rather than anchored to the semantic content specified by the part embeddings, leading to inconsistent multi-view outputs. Penalizing this feature variance directly improves view consistency.
We achieve this by extracting cross-attention feature maps $F$ from multiple layers of the network, which directly influence image content.
To sample an unseen part $m$, we generate a latent embedding from a Gaussian distribution:
\begin{equation} \label{eq:sampling}
    \begin{aligned}
    \widetilde{\mathbf{p}}_m \sim \mathcal{N}(\bm{\mu}_{m},\bm{\sigma}^2_{m}),
    \end{aligned}
\end{equation}
where $\bm{\mu}_{m} = \frac{1}{C} \sum \mathbf{p}^c_m$ and $\bm{\sigma}^2_{m} = \frac{1}{C} \sum (\mathbf{p}^c_m - \bm{\mu}_m)^2$ (\eg, $C=200$ for CUB-200 \cite{wah2011caltech}).
The sampled latent $\widetilde{\mathbf{p}}_{m}$ is then passed through $g$ to produce a textual embedding $\widetilde{\bm{t}}_{m}$.
The consistency loss $\mathcal{L}_{cl}$ is applied to minimize the L2-distance between feature maps across different noise instances:
\begin{align} \label{eq:cl}
    \mathcal{L}_{\text{cl}} = \mathbb{E}_{t,\epsilon,L} \left[ \sum_{i,j} \Vert F_{\epsilon_i}- F_{\epsilon_j}\Vert^2 \right],
\end{align}
where $i,j$ represent different random noise inputs at the same timestep $t$, and $L$ denotes different layers in the U-Net. In this way, the model is encouraged to produce similar internal representations for different noise instances at the same timestep $t$, promoting view-consistent generations. See Figure \ref{fig:consistency_loss} for an illustration.

\noindent
\textbf{Final objective function.}
The full optimization objective for \ours{} is:
\begin{align}
    \mathcal{L} = \mathcal{L}_{\text{diff}} + \lambda_{\text{attn}} \mathcal{L}_{\text{attn}} + \lambda_{\text{cl}} \mathcal{L}_{\text{cl}} + \lambda_{\text{reg}} \mathcal{L}_{\text{reg}},
\end{align}
where $\lambda_{\text{attn}}=0.01$, $\lambda_{\text{cl}}=0.001$, and $\lambda_{\text{reg}}=0.0001$ control the relative contributions of each loss term.

\begin{figure}[t]
    \centering
    \includegraphics[width=\linewidth]{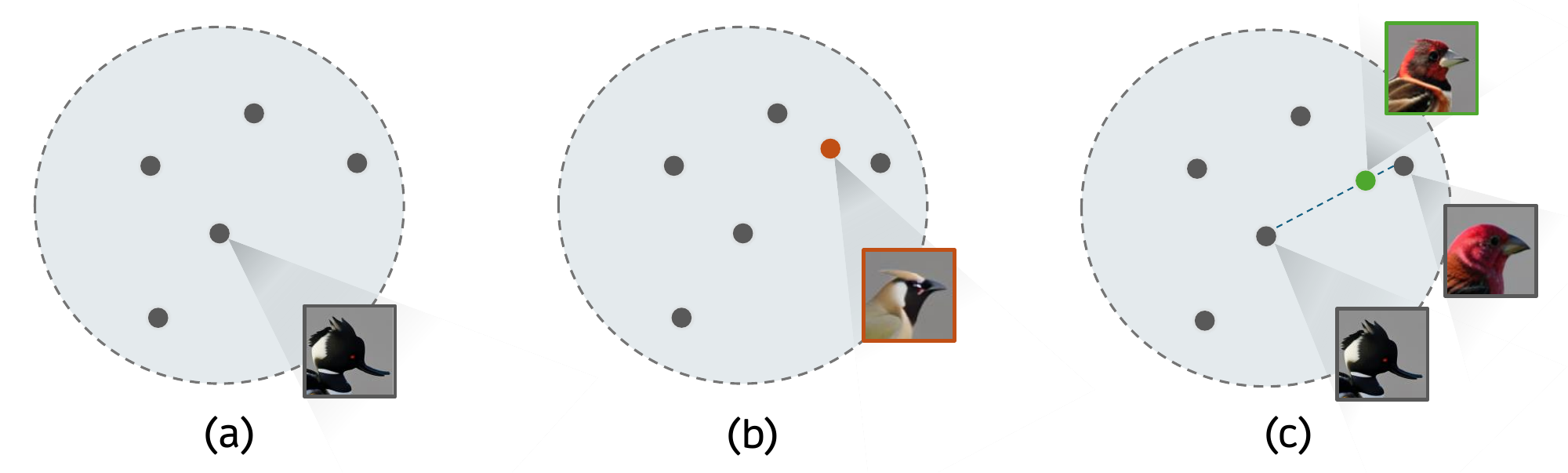}
    \caption{\textbf{(a)} Seen part selection generation. Unseen part synthesis via \textbf{(b)} novel sampling and \textbf{(c)} interpolation.
    }
    \vspace{-0.3cm}
    \label{fig:interpolation_illustration}
\end{figure}

\subsection{Creative object generation} \label{sec:creative_object_generation}

\noindent \textbf{Part-aware multi-view image generation.}
\ours{} allows users to compose new objects from existing or novel parts. We support three primary approaches:

\begin{enumerate}[(i)]
\item \textit{Seen part selection generation} (see Figure~\ref{fig:interpolation_illustration}(a)) allows users to compose new objects by selecting existing parts from seen species. Users can specify a text prompt such as ``a $\bm{t}^{*}_1$, ..., $\bm{t}^{*}_M$ bird", where each part is selected from different species, denoted by $*$.
\item \textit{Unseen part synthesis via novel sampling} (see Figure~\ref{fig:interpolation_illustration}(b)) creates new hybrid objects by directly sampling novel parts from the learned part latent space. This approach uses text prompts like ``a $\widetilde{\bm{t}}_1$, ..., $\widetilde{\bm{t}}_M$ bird", where each part $\widetilde{\bm{t}}_m$ is sampled via \eqref{eq:sampling}. 
\item \textit{Unseen part synthesis via interpolation} (see Figure~\ref{fig:interpolation_illustration}(c)) generates unique hybrids by interpolating part latents from different species. Specifically, a new part latent is computed as:
\begin{align}
    \hat{\bm{t}}_m=g([a \cdot \mathbf{p}^{c_1}_m + (1-a) \cdot \mathbf{p}^{c_2}_m ,\text{PE}_m])
\end{align}
where $\hat{\bm{t}}_m$ denotes the decoded textual embedding of the interpolated part latent, and $a$ controls the interpolation weight between species $c_1$ and species $c_2$.
\end{enumerate}

\noindent \textbf{3D generation.}
\ours{} supports 3D object generation through Score Distillation Sampling (SDS) \cite{poole2022dreamfusion}\footnote{We refer readers to the original paper for further details.}, compatible with the three previously defined approaches.
Unlike general text-to-3D generation, where broad prompts like ``bird'' lead to inconsistent results (see Appendix \ref{sec:lowgs_analysis}), each textual token in \ours{} corresponds to a specific part from a particular species, resulting in highly consistent appearance across different noise inputs (see Figure~\ref{fig:consistency_visual_comparison}(b)). This allows SDS optimization with a lower guidance scale, reducing oversaturation (see Figure~\ref{fig:3dgen}).
Our method works with both NeRF \cite{mildenhall2020nerf, poole2022dreamfusion} and 3DGS \cite{tang2024dreamgaussian, kerbl20233dgs}, and any improved 3D backend can be readily substituted.

%% file: pdfs/fig3-regloss.tex
\begin{figure}
    \centering
    \includegraphics[width=\linewidth]{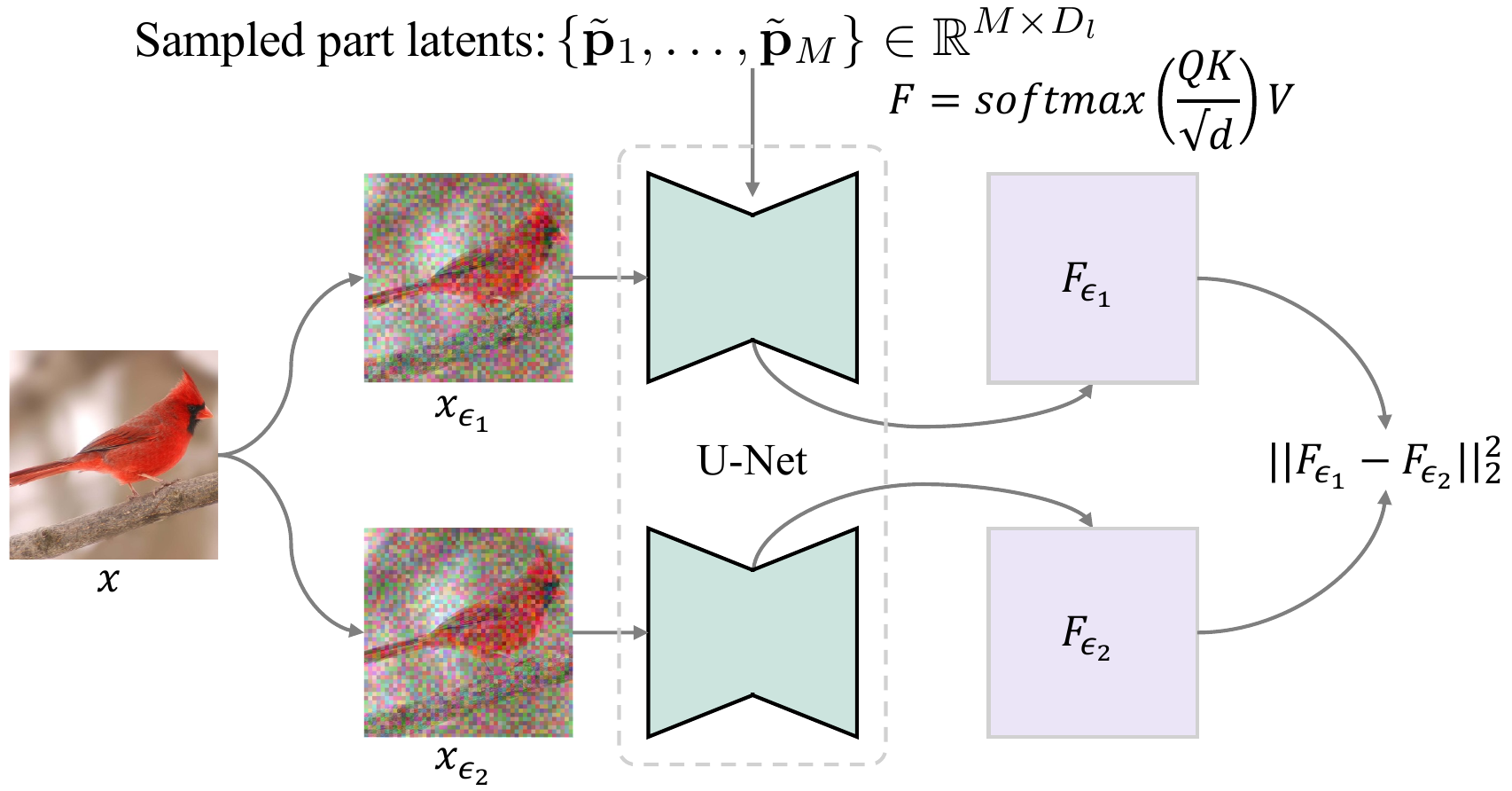}
    \caption{Illustration of the self-supervised feature consistency loss $\mathcal{L}_{\text{cl}}$. Given a part composition (seen or sampled), we apply two different noise instances at the same timestep and extract cross-attention feature maps $F$ from each. Minimizing the discrepancy between the two feature maps encourages the model to produce stable internal representations regardless of the specific noise realization, improving view consistency for both seen and unseen part compositions.
    }
    \vspace{-0.3cm} 
    \label{fig:consistency_loss}
\end{figure}

%% file: secs/3_exp.tex
\section{Experiment}

{\bf Datasets.}  
We conduct main experiments on the training set of fine-grained bird dataset CUB-200-2011~\cite{wah2011caltech}, which includes 200 bird species, with approximately 30 images per species.

\noindent{\bf Implementation.}  
All experiments were conducted on a single RTX 2080ti GPU ($\sim$8GB usage). 
We incorporate LoRA~\cite{hu2022lora} into the cross-attention layers of the diffusion backbone of MVDream~\cite{shi2023mvdream}, and only optimize LoRA parameters when training on single-view images. 
We apply attention loss on cross-attention maps at resolutions of $8\times8$ and $16\times16$.
We train the model for 100 epochs using batch size of 4 with gradient accumulation of 4 steps (effective batch size as 16) and a learning rate of 0.0001. For 3D object generation, we adopt threestudio~\cite{threestudio2023} as our framework. Our setup includes a species embedding dimension of $D_s = 768$, a part embedding dimension of $D_p = 4$\footnote{We also ablate with $D_p=16,32,64$ in the supplementary material.}, and a text embedding dimension of $D_t = 1024$. 

\noindent \textbf{Competitors.}
We compare \ours{} with two alternative models:
(a) \textit{Textual Inversion}: An adaptation of~\cite{gal2022textualinversion}, which learns a set of part-specific textual embeddings. Specifically, it optimizes $C \times M$ learnable word embeddings, where each embedding is defined as $\mathbf{t}^c_m \in \mathbb{R}^{D_t}$.
(b) \textit{PartCraft}: Extends (a) by incorporating a non-linear projector to further refine the learned word embeddings.
For a fair comparison, all models are implemented on MVDream with rank-4 LoRA layers~\cite{hu2022lora} and use attention loss $\mathcal{L}_{\text{attn}}$ (\eqref{eq:attn}) to enforce part disentanglement.

We compare against methods that operate under the same training setting (unposed single-view 2D images, no 3D data). Recent 3D-native approaches such as Part123~\cite{liu2024part123} and PartGen~\cite{chen2025partgen} require either posed multi-view images, 3D training data, or explicit 3D supervision, making direct comparison infeasible under our data-constrained setting.
The loss weight $\lambda_{\text{cl}}$ is ablated in the supplementary material.

\subsection{Multi-view generation evaluation}
We assess the quality of multi-view image generation by computing the average pairwise cosine similarity between the CLIP~\cite{radford2021clip} \& DINO~\cite{caron2021emerging} embeddings of generated multi-view images (treated as four separate images) and real species-specific images, following~\cite{ruiz2023dreambooth}. 

\input{tab_fig/tab_quan_scores}

\input{tab_fig/tab_partcomp}

\input{tab_fig/partcomp}

Figure~\ref{fig:appendix_clscompare} in Appendix~\ref{sec:appendix_cls} shows that both PartCraft and \ours{} effectively reconstruct the subject, whereas Textual Inversion struggles to preserve fine-grained details.
The subject fidelity metrics in Table~\ref{tab:quan_scores} confirm comparable fidelity between PartCraft and \ours{}, with \ours{} performing better on DINO similarity overall. We also evaluate FID and $\text{FID}_\text{CLIP}$, finding that \ours{} surpasses PartCraft by around 4\% in $\text{FID}_\text{CLIP}$.

\noindent \textbf{Part composition.}
To evaluate part disentanglement, we perform a \textit{part composition} experiment, swapping a specific part embedding with the corresponding embedding from a different species. Table~\ref{tab:partcomp_quan} presents a quantitative comparison with \textit{Exact Matching Rate (EMR)} and \textit{cosine similarity (CoSim)} from~\cite{ng2024partcraft}. For each of 1000 pairs, we generate 5 variations by swapping one part at a time. EMR checks if the generated part’s DINO feature is closest to the real class prototype; CoSim measures cosine similarity to the same prototype. See visual comparisons in Figure~\ref{fig:partcomp}. 

Among the methods, Textual Inversion exhibits the weakest performance, while PartCraft and \ours{} perform comparably. We attribute this to the shared projector used by both PartCraft and \ours{}, which maps part representations into textual embeddings through a learned function rather than optimizing them independently as in Textual Inversion.

\noindent \textbf{Linear interpolation.}   We analyze the latent space by interpolating between latent codes and progressively generating objects from left to right, as shown in Figure~\ref{fig:interpolate}. This serves as a probe for the continuity and expressiveness of each method's latent space:

\begin{enumerate}[(i)]
    \item \textit{Textual Inversion} suffers from unnatural transitions, as independently optimized word embeddings produce out-of-distribution results when interpolated.
    \item \textit{PartCraft} avoids major artifacts due to its embedding projector but exhibits abrupt switching, where a sudden change occurs at a particular interpolation step.
    \item \textit{\ours{}} shows more gradual and semantically coherent transitions, encouraged by our regularization $\mathcal{L}_{\text{reg}}$. This interpolation operates at the semantic part level, not low-level pixel or texture transitions.

\end{enumerate}

\noindent \textbf{Random sampling.}  
To evaluate the \textit{creativity} of different methods, we sample from the part latent space, compose novel objects, and compute two diversity measures: \textit{entropy ($H$)} and the \textit{effective number of classes ($e^{H}$)}. These metrics compare generated samples against training data to assess how broadly or narrowly the generated objects resemble real-world species. 

\input{tab_fig/interpolate}

\input{tab_fig/fig-consistency_visual_comparison}

\input{tab_fig/tab_us}

\input{tab_fig/random}

\input{tab_fig/tsne}

We extract DINO~\cite{caron2021emerging} feature embeddings from each generated multi-view image and retrieve the top-5 most similar species from the training set for each view. We then compute the frequency distribution of retrieved species.
To quantify diversity, we compute the \textit{entropy} of these frequencies and the \textit{effective number of classes ($e^{H}=\exp(H)$)}, which estimates the number of classes under a uniform distribution.
As shown in Table~\ref{tab:retrieval_evaluation}, \ours{} achieves a higher effective class count than PartCraft, indicating greater diversity in generating novel species. 

\input{tab_fig/tab_retrival}

Qualitative results in Figures~\ref{fig:random} and~\ref{fig:tsne} further support these findings: Textual Inversion frequently produces artifacts, PartCraft generates less diverse outputs, whereas \ours{} yields well-clustered DINO feature distributions and a broader range of diverse outputs.

\input{tab_fig/3dgen}

\input{pdfs/fig13-interpolation-sampling}

\subsection{3D generation evaluation} \label{sec:3dgen_evaluation}

We present NeRF-based\footnote{See supplementary material for 3DGS-based generation and image-to-3D methods~\cite{xu2024instantmesh}.} subject generation, novel generation (via random sampling), and part composition in Figure~\ref{fig:3dgen}. All generations are optimized using a standard guidance scale (\ie, $7.5$). Each SDS-based 3D object requires approximately 5 minutes of optimization on a single GPU.

Due to the lower guidance scale, PartCraft and Textual Inversion struggle to optimize high-quality 3D objects for novel generation and part composition, often producing artifacts such as overly smooth textures and small object outputs\footnote{Increasing the guidance scale can improve 3D object quality (\eg, generating larger outputs), but may introduce oversaturation artifacts.}.  
Our \ours{}, when trained with the feature consistency loss $\mathcal{L}_{\text{cl}}$, produces higher-quality 3D objects compared to the version without it (see Figure~\ref{fig:3dgen}), highlighting the importance of enforcing visual coherence across views.

We further compare against directly prompting MVDream with fine-grained species names (e.g., `a cardinal, 3d asset'). As shown in Appendix~\ref{sec:lowgs_analysis}, text prompts alone fail to produce consistent part-level details across seeds, as natural language lacks the granularity for subtle geometric and textural differences. In contrast, Chirpy3D's learned part latent codes capture structural variation that text cannot encode, and enable compositional operations (swapping, interpolation) impossible with text prompts.

\subsection{Further Analysis}

\noindent \textbf{Effect of $\mathcal{L}_{\text{cl}}$.}
Since there are no training examples for unseen random samples, the model faces a challenge in generating realistic outputs from unseen embeddings. Figure~\ref{fig:consistency_visual_comparison} compares our model with and without the proposed feature consistency loss $\mathcal{L}_{\text{cl}}$.

To further validate its impact, we conducted a user study: we generated 100 images for 10 existing and 10 randomly sampled species using models with and without $\mathcal{L}_{\text{cl}}$, and asked 20 users to vote for their preferred images. As shown in Table~\ref{tab:consistency_userstudy}, 82.5\% of users favored images generated with $\mathcal{L}_{\text{cl}}$, confirming that the loss meaningfully improves visual coherence.

%% file: tab_fig/tab_quan_scores.tex
\begin{table}[t]
    \centering
    \adjustbox{max width=\linewidth}{
        \begin{tabular}{l|cccc}
        Method & DINO $\uparrow$ & CLIP $\uparrow$ & FID $\downarrow$ & $\text{FID}_\text{CLIP}$ $\downarrow$\\
             \toprule
        \TI{}~\cite{gal2022textualinversion} & 0.357 & 0.594 & 43.86 & 30.89 \\
        \PC{}~\cite{ng2024partcraft} & 0.365 & 0.597 & \bf 43.25 & 28.98 \\
        \bf \method{} & \bf 0.380 & \bf 0.619 & 43.41 &
        \bf 27.84 \\
        \end{tabular}
    }
    \vspace{-2mm}
    \caption{
    Results in subject fidelity metrics.}
    \vspace{-2mm}
    \label{tab:quan_scores}
\end{table}

%% file: tab_fig/tab_partcomp.tex
\begin{table}[t]
    \centering
    \adjustbox{max width=\linewidth}{
    \begin{tabular}{l|cc}
         Method &  EMR $\uparrow$ & CoSim $\uparrow$ \\
         \toprule
        Textual Inversion~\cite{gal2022textualinversion}  & 0.210 & 0.691 \\ 
        PartCraft~\cite{ng2024partcraft} & 0.291 & 0.722 \\
        \bf \ours{} & \bf 0.295 & \bf 0.724
    \end{tabular}
    }
    \vspace{-2mm}
    \caption{Part composition results.}
    \vspace{-0.4cm}
    \label{tab:partcomp_quan}
\end{table}

%% file: tab_fig/partcomp.tex
\begin{figure*}[t]
    \centering
        \includegraphics[width=\linewidth]{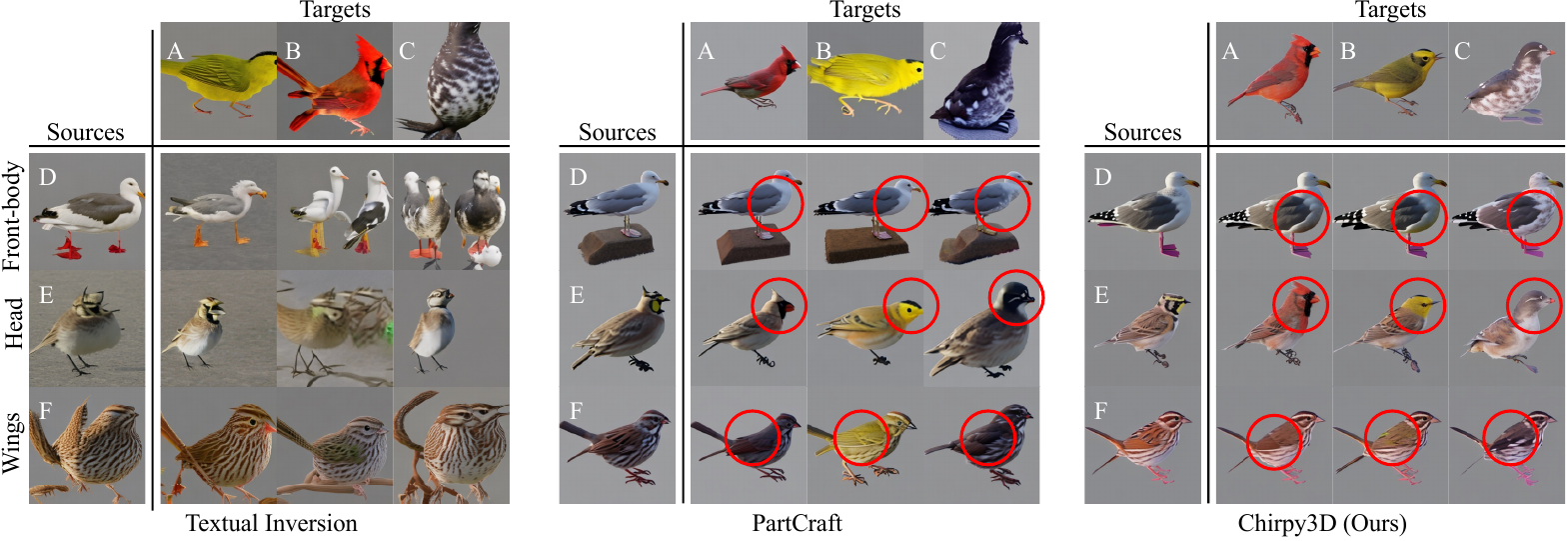}
    \caption{
    Each image is generated from text prompts like ``a $\mathbf{t}_1,...,\mathbf{t}_M$ bird''. Each source/target image is generated using all parts from one species. For each row, we swap one part embedding (\eg, head) with that from target species.
    All images are generated independently from text prompts, so slight global variations such as changes in texture may appear. This is due to the holistic nature of diffusion denoising.
    }
    \vspace{-0.1cm}
    \label{fig:partcomp}
\end{figure*}

%% file: tab_fig/interpolate.tex
\begin{figure}[t]
    \centering
    \begin{subfigure}[b]{\linewidth}
        \centering
        \includegraphics[width=\linewidth]{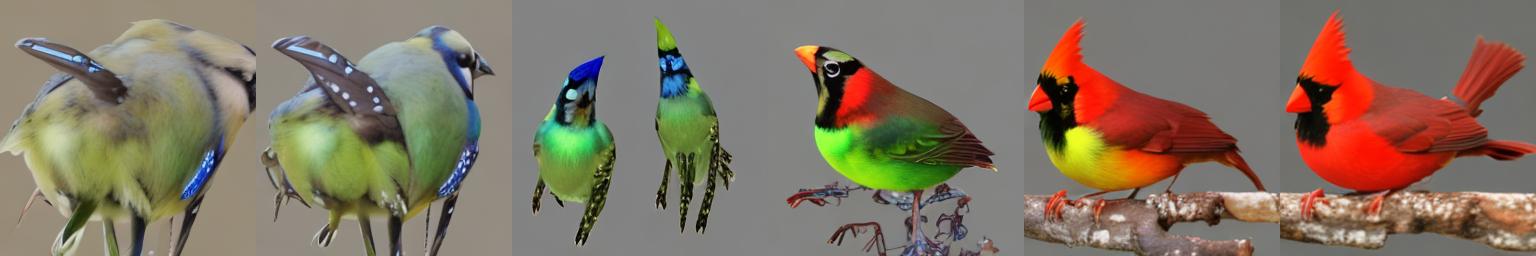}
        \caption{Textual Inversion}
    \end{subfigure}
    \begin{subfigure}[b]{\linewidth}
        \centering
        \includegraphics[width=\linewidth]{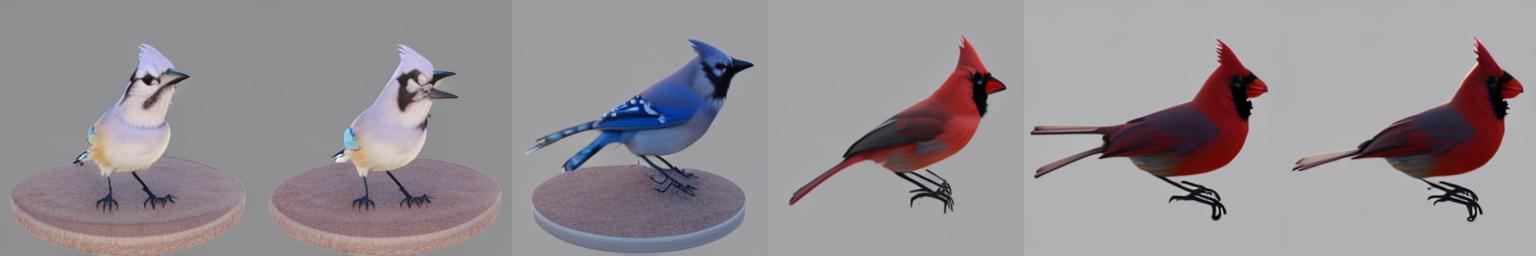}
        \caption{PartCraft}
    \end{subfigure}
    \begin{subfigure}[b]{\linewidth}
        \centering
        \includegraphics[width=\linewidth]{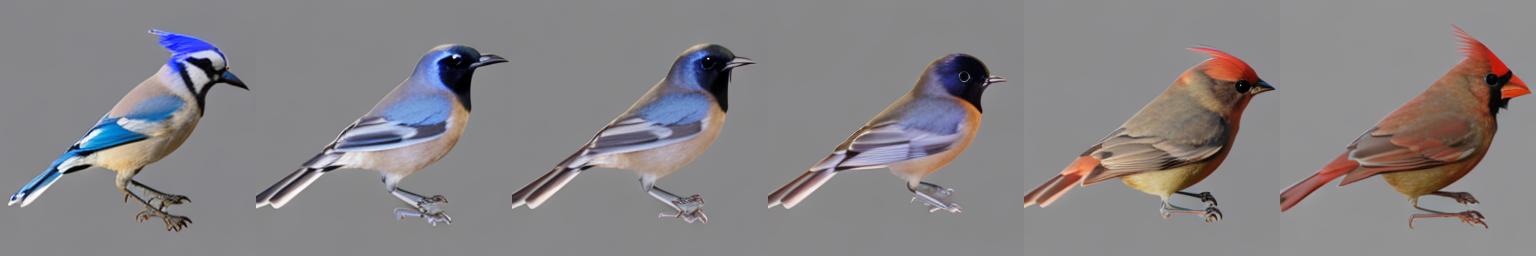}
        \caption{\ours{} (Ours)}
    \end{subfigure}
    \vspace{-6mm}
    \caption{
    Linear interpolation of all part latents between two different species -- \textit{blue jay} and \textit{cardinal}. 
    Only one view is shown. 
    Our \ours{} enables more semantically consistent and gradual transitions, while PartCraft exhibits abrupt semantic switches mid-way. See more at Figure~\ref{fig:appendix_interpolationsample} in Appendix~\ref{sec:appendix_interpolation}.
    }
    \label{fig:interpolate}
\end{figure}

%% file: tab_fig/fig-consistency_visual_comparison.tex
\begin{figure}[t]
    \centering
    \begin{subfigure}[b]{0.49\linewidth}
        \includegraphics[width=\linewidth]{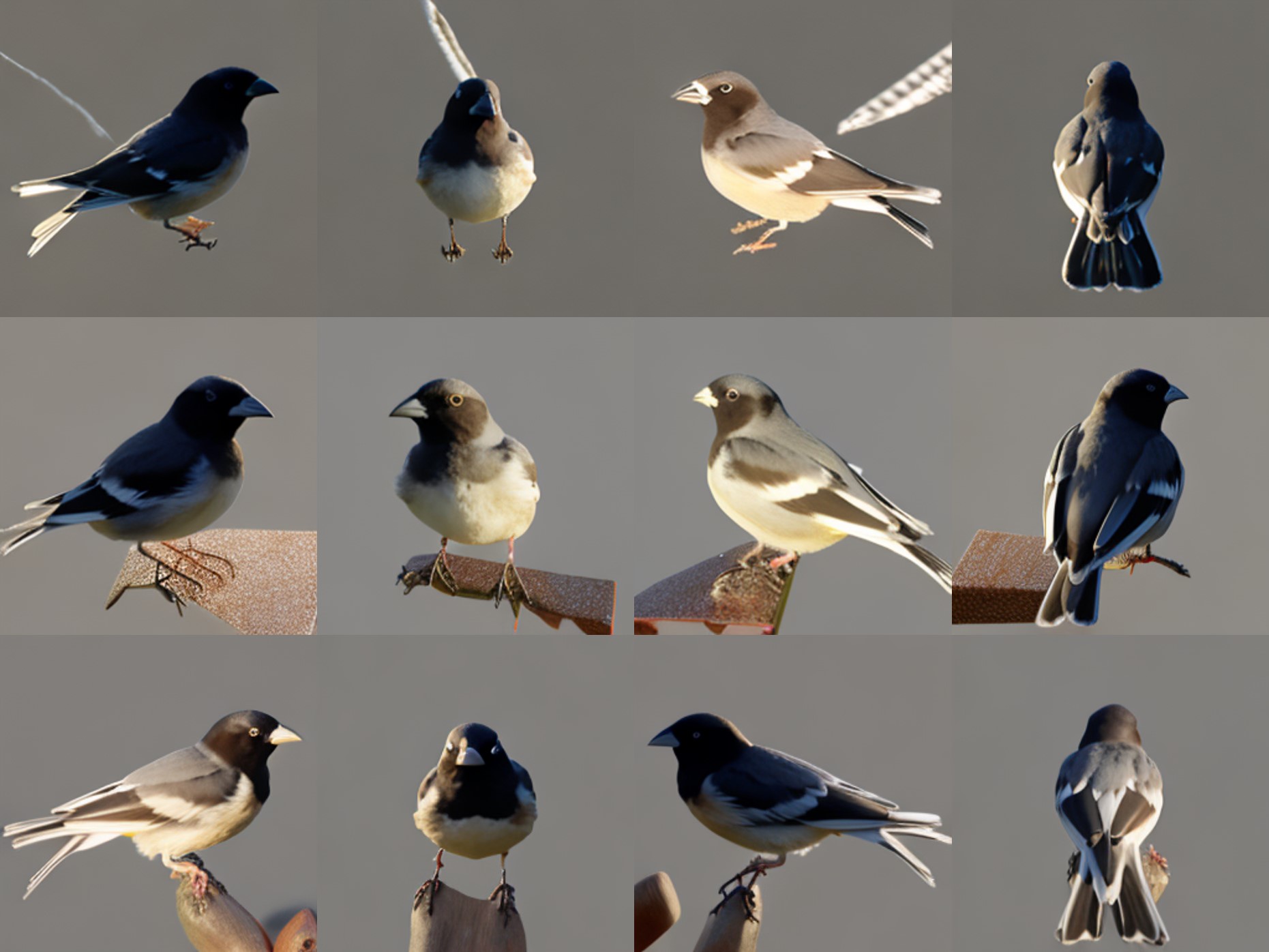}
        \caption{Ours without $\mathcal{L}_{\text{cl}}$}
    \end{subfigure} \hfill
    \begin{subfigure}[b]{0.49\linewidth}
        \includegraphics[width=\linewidth]{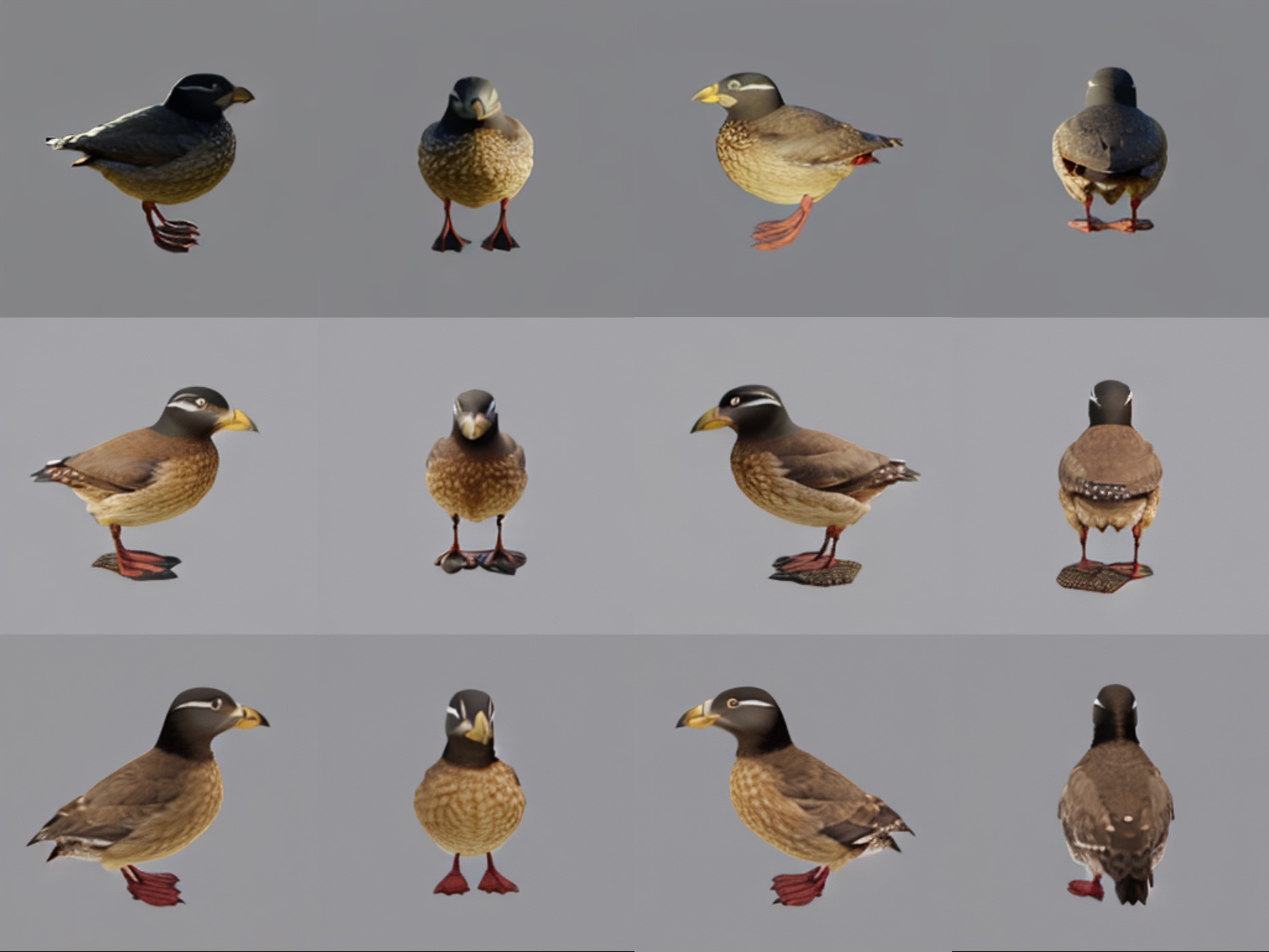}
        \caption{Ours with $\mathcal{L}_{\text{cl}}$}
    \end{subfigure}
    \vspace{-2mm}
    \caption{
    All images are generated with the same camera pose but with different seeds on \textit{unseen} (novel) latent. (a) Without our feature consistency loss $\mathcal{L}_{\text{cl}}$, the generated images exhibit more artifacts and inconsistent visual features compared to (b). See more at Figure~\ref{fig:appendix_after_cl} in Appendix~\ref{sec:appendix_beforeafter_cl}.
    }
    \vspace{-3mm}
    \label{fig:consistency_visual_comparison}
\end{figure}

%% file: tab_fig/tab_us.tex
\begin{table}[t]
    \centering
    \adjustbox{max width=\linewidth}{
        \begin{tabular}{l|cc}
            \bf \ours{} & With $\mathcal{L}_{\text{cl}}$ & Without $\mathcal{L}_{\text{cl}}$ \\
             \midrule
            Preference $\uparrow$ & \bf 82.5\% & 17.5\%
        \end{tabular}
    }
    \caption{User study to verify $\mathcal{L}_{\text{cl}}$.
    }
    \label{tab:consistency_userstudy}
\end{table}

%% file: tab_fig/random.tex
\begin{figure*}[t]
    \centering
    \begin{subfigure}[b]{0.32\textwidth}
        \centering
        \includegraphics[height=0.78\linewidth]{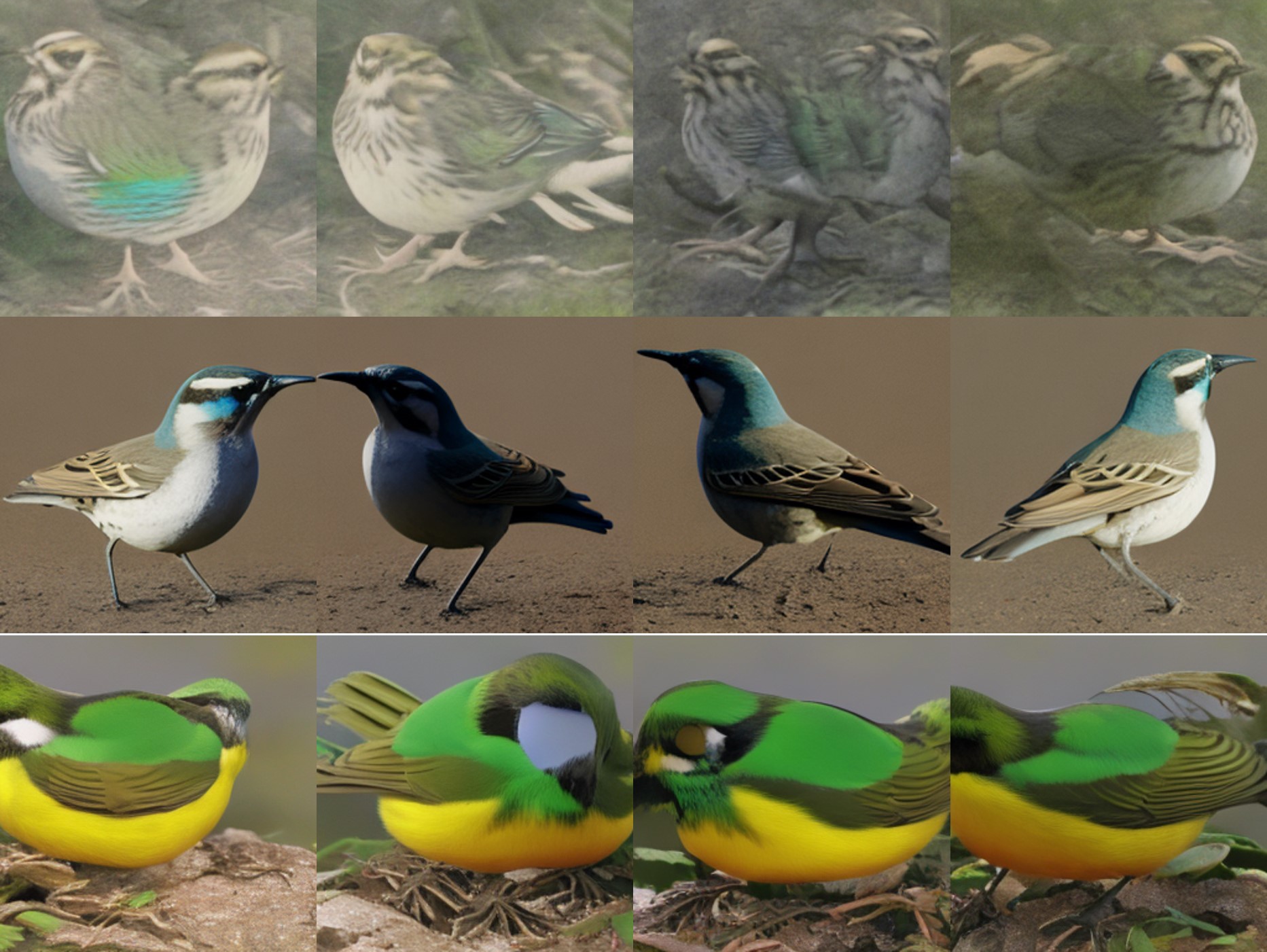}
        \caption{Textual Inversion}
    \end{subfigure} \hfill
    \begin{subfigure}[b]{0.32\textwidth}
        \centering
        \includegraphics[height=0.78\linewidth]{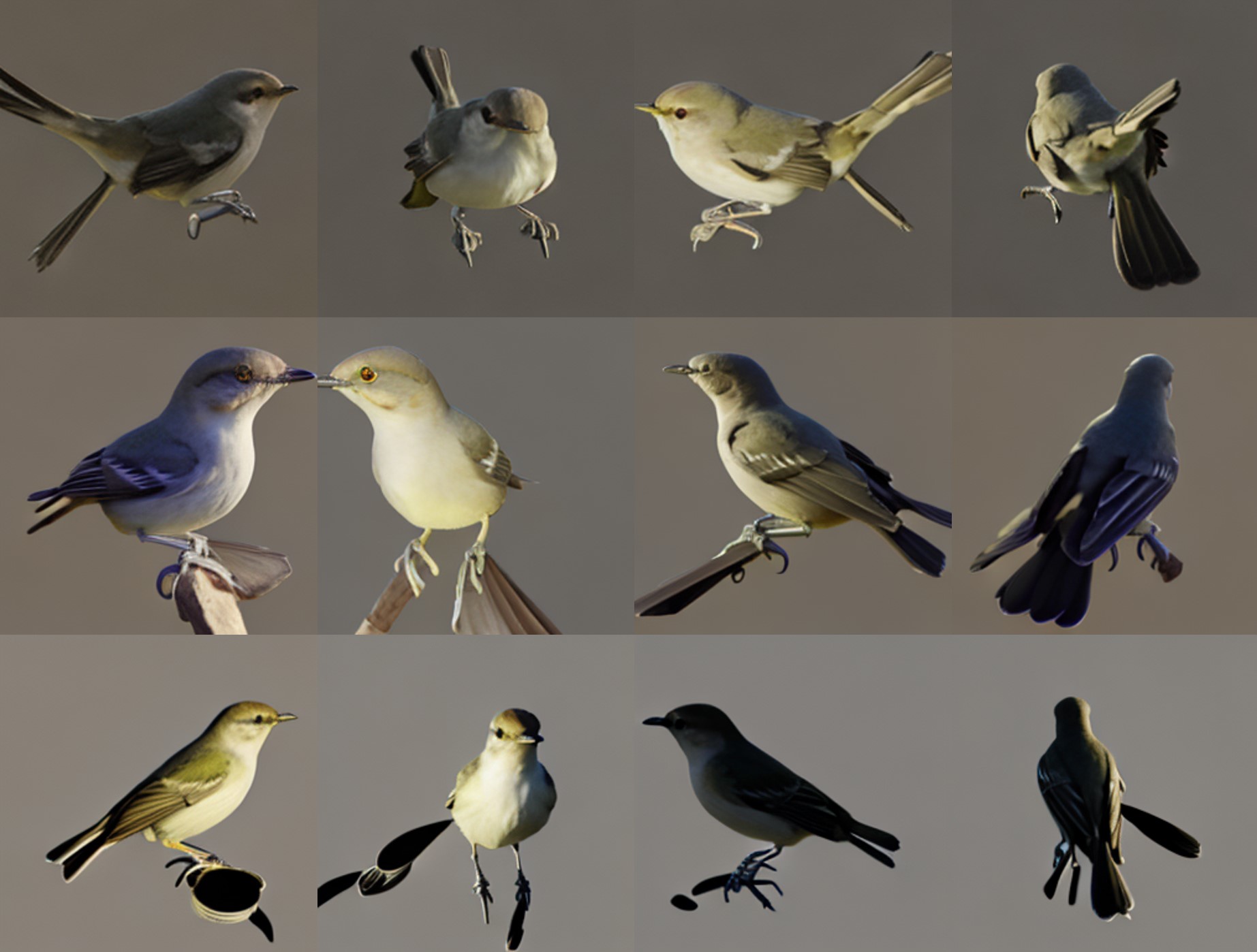}
        \caption{PartCraft}
    \end{subfigure}
    \hfill
    \begin{subfigure}[b]{0.32\textwidth}
        \centering
        \includegraphics[height=0.78\linewidth]{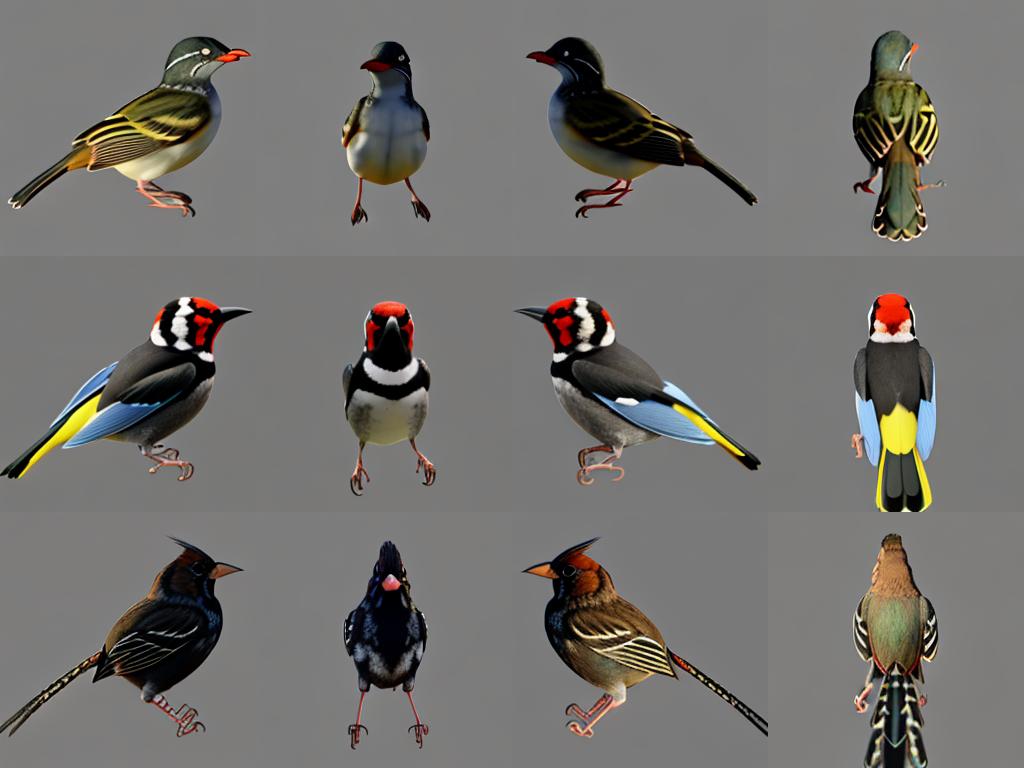}
        \caption{\ours{} (Ours)}
    \end{subfigure}
    \caption{Generated images with random sampled latents/embeddings. \TI{} often produces artifacts due to out-of-distribution word embeddings. \PC{} generates images with fewer artifacts but lacks consistency. In contrast, our \ours{} generates novel images with greater diversity. See more at Figure~\ref{fig:appendix_randomcompare} in Appendix~\ref{sec:appendix_randomsampling}.
    }
    \label{fig:random}
\end{figure*}

%% file: tab_fig/tsne.tex
\begin{figure}[t]
    \centering
    \begin{subfigure}[b]{0.32\linewidth}
        \centering
        \includegraphics[width=\linewidth]{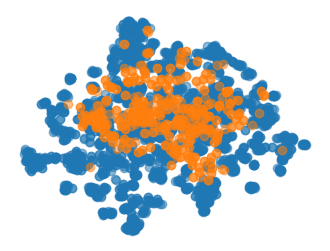}
        \caption{Textual Inversion}
    \end{subfigure} \hfill
    \begin{subfigure}[b]{0.32\linewidth}
        \centering
        \includegraphics[width=\linewidth]{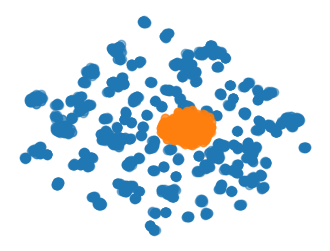}
        \caption{PartCraft}
    \end{subfigure} \hfill
    \begin{subfigure}[b]{0.32\linewidth}
        \centering
        \includegraphics[width=\linewidth]{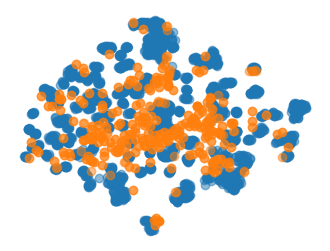}
        \caption{\ours{} (Ours)}
    \end{subfigure}
    \vspace{-2mm}
    \caption{t-SNE embeddings of DINO features of generated images using \TI{}, \PC{} and \ours{}. \textcolor{blue}{Blue} represents generated images of seen classes; \textcolor{orange}{Orange} represents images of novel generation.
    }
    \label{fig:tsne}
\end{figure}

%% file: tab_fig/tab_retrival.tex
\begin{table}[t]
    \centering
    \adjustbox{max width=\linewidth}{
        \begin{tabular}{l|cc}
            Method & $H$ $\uparrow$ & $e^H$ $\uparrow$ \\
             \toprule
            PartCraft~\cite{ng2024partcraft} & 4.07 & 58.6 \\
            \bf \ours{} & 4.81 & 123.2 \\
        \end{tabular}
    }

    \caption{Diversity test. Higher values indicate greater diversity.}
    \vspace{-0.3cm}
    \label{tab:retrieval_evaluation}
\end{table}

%% file: tab_fig/3dgen.tex
\begin{figure}[t]
    \centering
    \includegraphics[width=\linewidth]{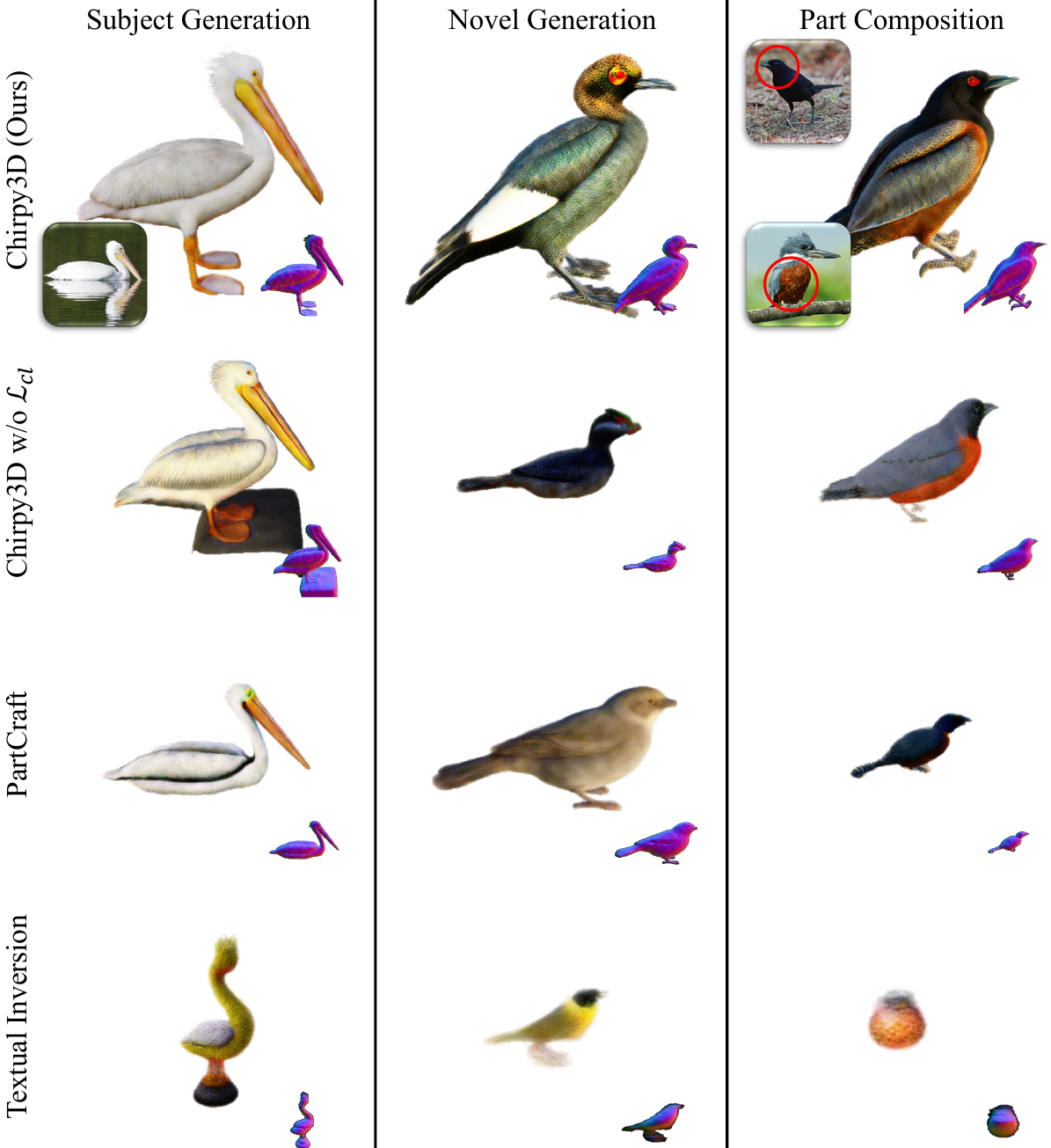}
    \caption{NeRF rendering of learned 3D objects.}
    \label{fig:3dgen}
\end{figure}

%% file: pdfs/fig13-interpolation-sampling.tex
\begin{figure*}[t]
    \centering
    \includegraphics[width=\linewidth]{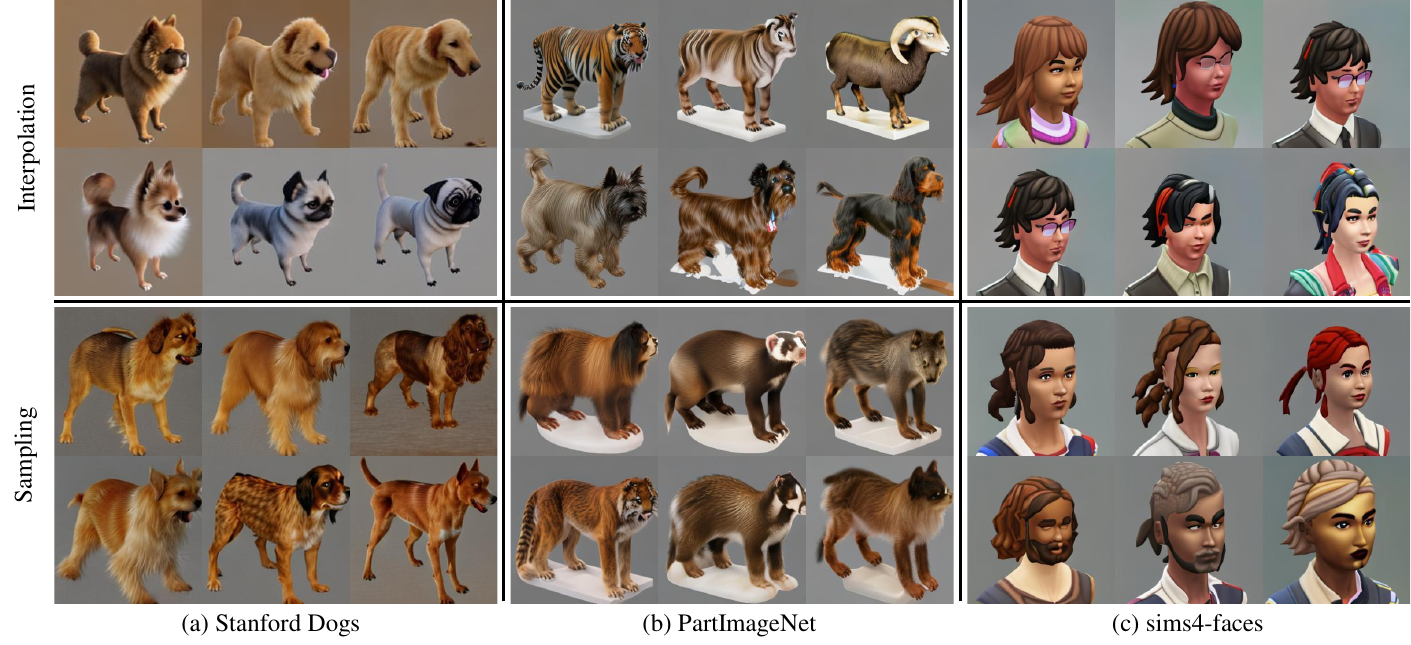}
    \vspace{-6mm}
    \caption{\textbf{(Top)} \textbf{Middle} images indicate a hybrid between species/instance \textbf{left} and \textbf{right} by linearly interpolating their part latent codes. \textbf{(Bottom)} Generated images with random sampled latents for PartImageNet (quadruped only) \cite{he2021partimagenet} and sims4-faces \cite{sims4faces2022}. We regularize part embeddings to follow a standard Gaussian distribution to encourage interpolation and sampling.
    }
    \vspace{-4mm}
    \label{fig:interpolation-sampling}
\end{figure*}

%% file: secs/4_conclusion.tex
\section{Conclusion}

We presented \ours{}, a framework that learns a part-aware multi-view diffusion model from unposed 2D images, enabling fine-grained, compositional object generation in a zero-shot setting. \ours{} decomposes each object into a set of semantic parts, learned without 3D supervision or manual part annotations, and models each part as a continuous distribution in latent space. This enables flexible recombination, structured interpolation, and compositional synthesis across species, including downstream 3D generation.

Our experiments demonstrate that \ours{} achieves competitive or improved performance over existing baselines in visual quality and fine-grained control. While this work primarily focuses on birds, \ours{} shows promising generalization to other fine-grained categories, including dogs, quadrupeds, and character faces (see Figure~\ref{fig:interpolation-sampling}), suggesting potential for applications in gaming and virtual content creation.

\noindent \textbf{Limitations and Future Work.}
\ours{} inherits limitations from its diffusion backbone, including multi-view inconsistencies and limited pose controllability, and the learned part latent codes may exhibit entanglement when mixing structurally incompatible parts.
The framework also assumes a fixed number of semantic parts ($M=5$) and relies on 2D part segmentation masks from an off-the-shelf detector~\cite{ng2024partcraft}, whose quality may affect disentanglement for occluded or ambiguous regions.
For 3D generation, SDS-based optimization requires per-object iterative refinement (${\sim}$5 min per object); however, our core contribution is the part-aware multi-view diffusion model, and improved 3D backends can be directly substituted.
Extending to variable part counts and diverse articulations is an important future direction.

%% file: supp.tex
\clearpage
\setcounter{page}{1}
\maketitlesupplementary
\appendix

\section{Video Results}

Please open the \texttt{index.html} file to view a 360-degree video showcasing examples of 3D birds. In case the HTML website is not displaying correctly, please see \texttt{assets/videos/*/*.mp4} for 3D birds video and \texttt{assets/additional\_videos/*/*.mp4} for additional examples.

\section{Derivation}
We use the symbol \( x \) to represent the part latent \( \mathbf{p} \) for clarity. Assuming a multivariate Gaussian distribution with a spherical covariance matrix \(\sigma^2 \mathbb{I}\), the probability density function is expressed as:
\begin{align}
    p(\bm{x}) = \frac{1}{\sqrt{2 \pi \sigma^2}} \exp{\bigg( - \frac{\Vert \bm{x} - \bm{\mu} \Vert^2}{2 \sigma^2} \bigg)}.
\end{align}
Since we assume a zero-mean Gaussian distribution, i.e., \(\bm{\mu} = 0\), the expression simplifies to:
\begin{align}
    p(\bm{x}) = \frac{1}{\sqrt{2 \pi \sigma^2}} \exp{\bigg( - \frac{\Vert \bm{x} \Vert^2}{2 \sigma^2} \bigg)}.
\end{align}
Taking the logarithm of \(p(\bm{x})\) yields:
\begin{align}
    \log p(\bm{x}) = \log \bigg( \frac{1}{\sqrt{2 \pi \sigma^2}} \bigg) + \bigg( - \frac{\Vert \bm{x} \Vert^2}{2 \sigma^2} \bigg)
\end{align}
Maximizing \(\log p(\bm{x})\) is therefore equivalent to minimizing the term:
\begin{align}
    \mathcal{L}_{\text{reg}} = \frac{\Vert \bm{x} \Vert^2}{2 \sigma^2}
\end{align}
In practice, we set \(\sigma^2 = 1\) for simplicity, leading to the regularization loss $\mathcal{L}_{\text{reg}}$.

\section{Implementation Details}

\subsection{Settings}
For finetuning MVDream, we select AdamW \cite{loshchilov2018adamw} as optimizer and learn at a constant learning rate of 0.0001 and weight decay of 0.01. 
We apply random horizontal flipping for data augmentation. 
Following MVDream~\cite{shi2023mvdream}, we train on images with resolution $256 \times 256$.
We integrate the LoRA design \cite{hu2022lora} from \texttt{diffusers} library\footnote{\url{https://github.com/huggingface/diffusers/blob/main/examples/text_to_image/train_text_to_image_lora.py}}, in which the low-rank adapters are added to the $QKV$ and $out$ components of all cross-attention modules.
For attention loss $\mathcal{L}_{\text{attn}}$, we select cross-attention maps with feature map sizes of $8 \times 8$ and $16 \times 16$, given input resolution of $256 \times 256$.
For 3D generation described in Section~\ref{sec:creative_object_generation} \& Section~\ref{sec:3dgen_evaluation}, including generated objects in Figure~\ref{fig:big-teaser} and supplementary material, we build upon \texttt{MVDream-threestudio}
\footnote{\url{https://github.com/bytedance/MVDream-threestudio}}. 
The SDS loss $\mathcal{L}_{\text{SDS}}$ used in our setup follows MVDream's \cite{shi2023mvdream} $x_0$-reconstruction loss, with a CFG rescale of $0.3$. 

\subsection{Competitors}
Table~\ref{tab:comparison} shows the comparison of different methods in components.
\input{tab_fig/ab}

\begin{algorithm}[t]
\caption{Code snippet for diversity evaluation}
\label{alg:ret}
\definecolor{codeblue}{rgb}{0.25,0.5,0.5}
\lstset{
  backgroundcolor=\color{white},
  basicstyle=\fontsize{7.2pt}{7.2pt}\ttfamily\selectfont,
  columns=fullflexible,
  breaklines=true,
  captionpos=b,
  commentstyle=\fontsize{7.2pt}{7.2pt}\color{gray},
  keywordstyle=\fontsize{7.2pt}{7.2pt}\color{codeblue},
}
\begin{lstlisting}[language=python]
# db_feats: DINO feats of training images (Nx768)
# query_feats: DINO feats of generated images (Qx768)
# db_labels: Training ground-truth labels (N)

def retrieve(query_features, db_features, db_labels):
    # both features assume normalized
    cossim = query_features @ db_features.t()  # (Q, N)
    indices = torch.argsort(cossim, dim=-1)
    top5 = indices[:, :5]
    idx = torch.arange(len(db_features)).to(db_labels.device)
    return (
        db_labels[top5.reshape(-1)].reshape(top5.size(0), top5.size(1))
    )
    
def histogram_entropy(hist):
    # Normalize the histogram to obtain probabilities
    prob = hist / hist.sum()
    
    # Compute the entropy
    entropy = -torch.sum(prob * torch.log(prob))
    
    return entropy

retrieve_labels = retrieve(
    query_feats, 
    db_feats, 
    db_labels
)

hist1 = torch.zeros(200)
for i in range(200):
    hist1[i] = (retrieve_labels == i).sum()
    
entropy_hist1 = histogram_entropy(hist1)
print(f"Entropy of {n}:", entropy_hist1.item())
print("Num classes", torch.exp(entropy_hist1).item())

\end{lstlisting}
\end{algorithm}

\subsection{Algorithms}
For the evaluation algorithm, please refer to Algorithm \ref{alg:ret}.

\section{Ablation Study}
\begin{figure}[h]
    \centering
    \begin{subfigure}[b]{0.32\linewidth}
        \centering
        \includegraphics[width=\linewidth]{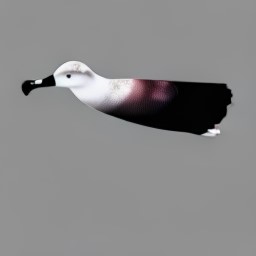}
        \caption{$0.1$}
    \end{subfigure} \hfill
    \begin{subfigure}[b]{0.32\linewidth}
        \centering
        \includegraphics[width=\linewidth]{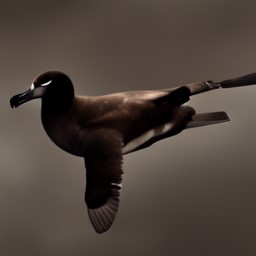}
        \caption{$0.01$}
    \end{subfigure} \hfill
    \begin{subfigure}[b]{0.32\linewidth}
        \centering
        \includegraphics[width=\linewidth]{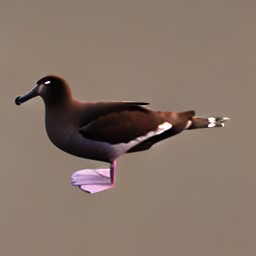}
        \caption{$0.001$}
    \end{subfigure}
    \caption{Comparison of generated images with different scales, $\lambda_{\text{cl}}$, for $\mathcal{L}_{\text{cl}}$.}
    \vspace{-4mm}
    \label{fig:losscl}
\end{figure}

\noindent \textbf{Weights for $\mathcal{L}_{\text{cl}}$.} We set $\lambda_{\text{cl}} \le 0.001$, as higher values degrade generation quality, often causing failures in object generation. See Figure \ref{fig:losscl}.

\noindent \textbf{Dimension of $D_p$.} Table \ref{tab:latentdimension} presents a comparison of different variants with varying $D_p$ dimensions. We choose $D_p=4$ as the default setting, as it achieves highest diversity in random sampling.
\begin{table}[h]
    \centering
    \begin{tabular}{c|cccc}
        $D_p$ & 4 & 16 & 32 & 64 \\
        \hline
        $H$ & \bf 4.81 & 4.68 & 4.67 & 4.33 \\
        $e^H$ & \bf 123.2 & 108.1 & 106.7 & 76.3 \\
    \end{tabular}
    \caption{Entropy $H$ and the effective number of classes $e^H$ of Top-5 retrieved classes using generated images from random part latents. Higher values indicate greater diversity.}
    \label{tab:latentdimension}
    \vspace{-4mm}
\end{table}

\section{Further Analysis}

\begin{figure}[h]
    \centering
    \includegraphics[width=0.5\linewidth]{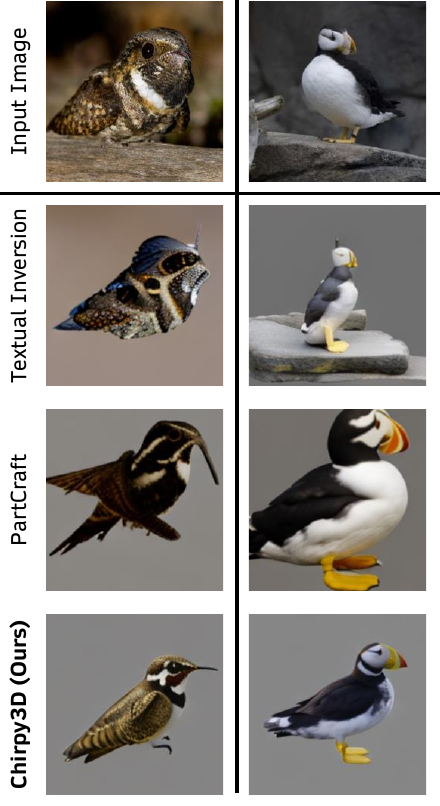}
    \caption{Image reconstruction from part latent inversion. We run 1,000 learning steps for optimization.}
    \label{fig:inverse_experiment}
    \vspace{-4mm}
\end{figure}

\noindent \textbf{Inversion experiment.}
Once trained, we can perform part code inversion on input images, and use the learned part code to reconstruct original images. 
Figure~\ref{fig:inverse_experiment} shows the visual comparison among three methods. Textual Inversion struggles with texture quality. PartCraft exhibits artifacts, possibly due to the ambiguity of input samples (e.g., camouflage birds). In contrast, our method successfully reconstructs the input with accurate size and significantly improved texture quality.

\begin{figure}[h]
    \centering
    \includegraphics[width=\linewidth]{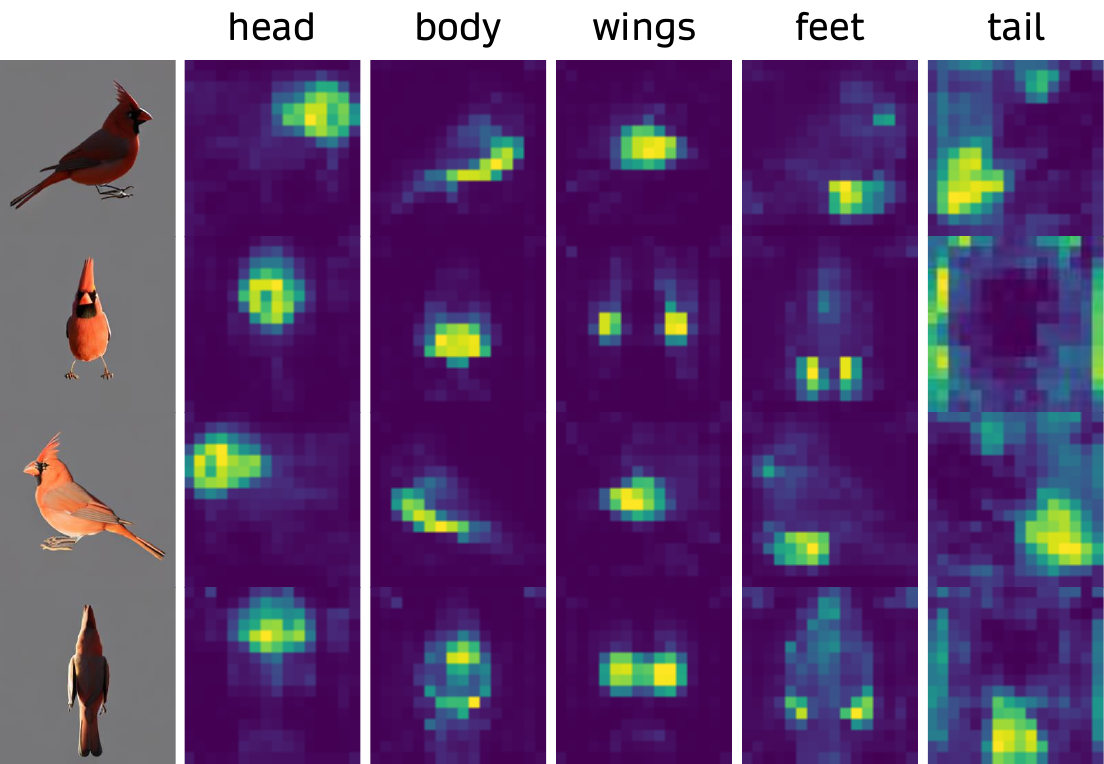}
    \caption{Visualization of cross-attention maps (averaged over all layers with size $16\times16$.) of multi-view images.}
    \label{fig:attentionmap}
\end{figure}

\noindent \textbf{Effect of $\mathcal{L}_{attn}$.}
Figure \ref{fig:attentionmap} shows the visualization of the averaged cross-attention maps from multi-view images. 
Although we fine-tune MVDream exclusively on 2D images, it successfully transfers to multi-view image generation, leveraging its learned prior. 
The cross-attention map highlights a strong correlation between the image and the tokens of the corresponding parts, demonstrating clear disentanglement -- each part avoids attending to incorrect spatial locations, thanks to the part attention loss. 
Furthermore, when parts of the image are occluded (e.g., the tail is occluded when facing forward), the model attends to the background rather than incorrectly focusing on other parts.

\begin{table}[h]
    \centering
    \adjustbox{max width=\linewidth}{
        \begin{tabular}{c|ccc}
             & Textual Inversion & PartCraft & \ours{} (Ours) \\
             \hline
             1 - IoU &  0.744 & 0.954 & \bf 0.957 \\
        \end{tabular}
    }
    \caption{Overlapping score between the cross-attention map of all parts.}
    \label{tab:attentionmap_overlap}
\end{table}

\noindent \textbf{Analysis of part disentanglement.}
In Table \ref{tab:attentionmap_overlap}, we present the overlap score, defined as \(1 - \text{IoU}\) (intersection-over-union), calculated between the cross-attention maps of all parts (see an example in Fig.~\ref{fig:attentionmap}). The scores are averaged over 1,000 samples. These results demonstrate that the parts are effectively disentangled during generation using our \ours{}, enabling part-aware generation. In contrast, Textual Inversion struggles with part-aware generation, possibly because the learned word embeddings for each part are independent and lack mutual awareness, making it difficult to generate a cohesive whole object from individual parts.

\begin{figure}[h]
    \centering
    \includegraphics[width=\linewidth]{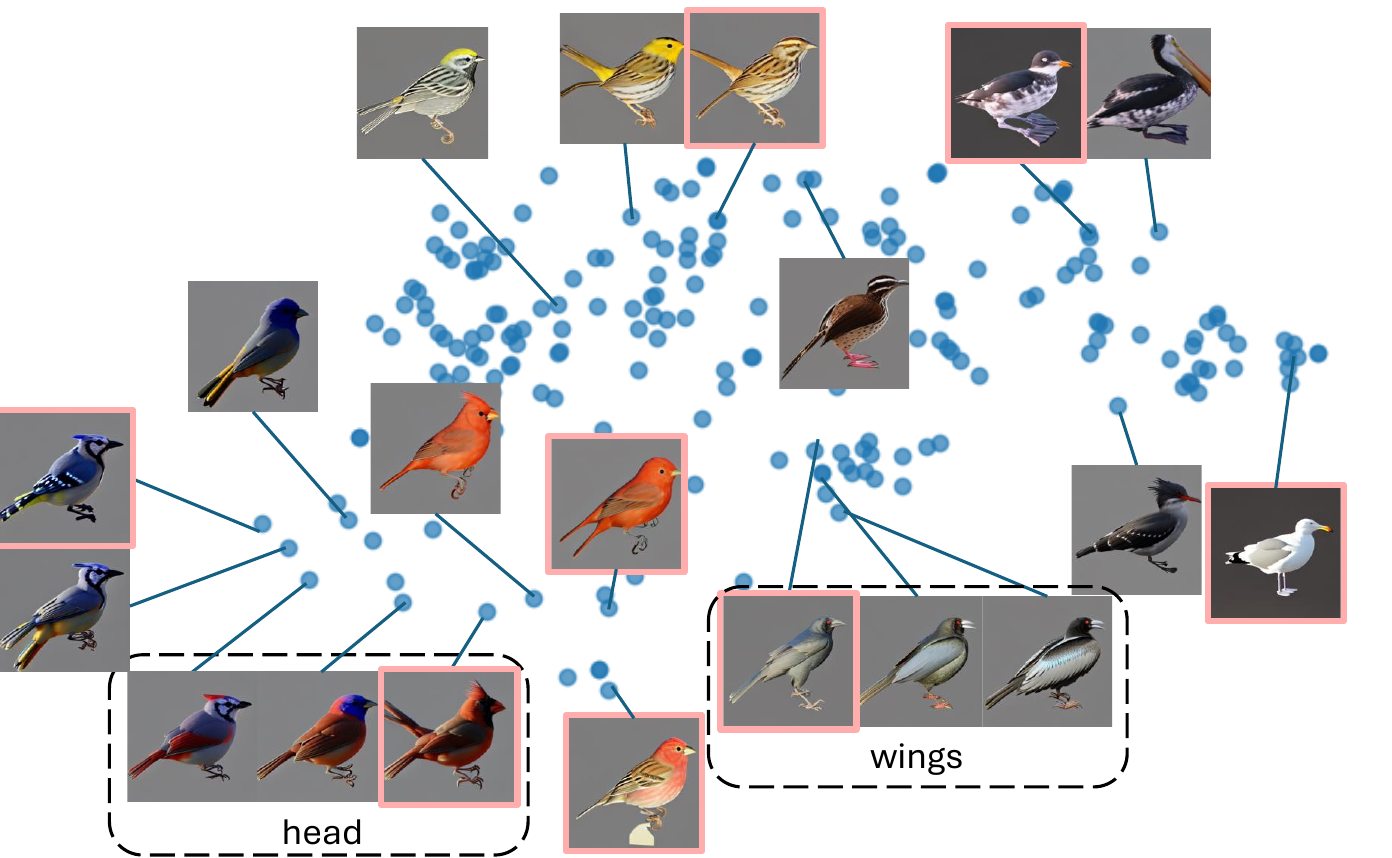}
    \caption{Visualizing part latent space via t-SNE embeddings.}
    \label{fig:partlatentexplore}
\end{figure}

We display the part latent space in Figure \ref{fig:partlatentexplore}. In this latent space, we can traverse to sample desired species/generations. For instance, we can perform interpolation between two species, randomly sample, and also perform part composition, all can be done within this part latent space.

\section{Different representations for 3D object Generation}

\subsection{NeRF recipe}

For NeRF-based 3D generation, we use \texttt{threestudio} \cite{threestudio2023} as our framework and \texttt{MVDream-threestudio} as a plugin. We implement a custom prompt processor to handle the input prompt, as we replace the word embeddings using Eq.~(3) in the main paper. After tokenization and before passing the word embeddings, we substitute the placeholder's word embedding with our computed word embedding, $\bm{t}$. An example prompt with placeholders is formatted as ``a [part$_1$] ... [part$_M$] bird.''

We use all default settings, except for setting the number of samples per ray to 256 to prevent out-of-memory errors and disabling background color augmentation. 

The guidance scale is set to $7.5$ by default. Different timesteps affect the SDS optimization process \cite{huang2023dreamtime}. Lower timesteps (less noise) emphasize detailed textures, while higher timesteps (more noise) focus on coarse structure. During SDS loss optimization, the timestep is randomly sampled within a specified minimum and maximum range. 

We set the minimum timestep range to decay from 0.98 to 0.02 over 3,000 steps and the maximum timestep range to decay from 0.98 to 0.3 over 8,000 steps. We observe that bird structures begin forming around 1,000 steps. Therefore, we quickly decay the minimum timesteps to 0.02 to allow the model to focus on texture details during the early stages of training.

\subsection{3DGS recipe}

For 3DGS-based generation, we use DreamGaussian's \cite{tang2024dreamgaussian} official implementation, with the training strategy used in NeRF recipe.

\begin{figure}[h]
    \centering
    \includegraphics[width=\linewidth]{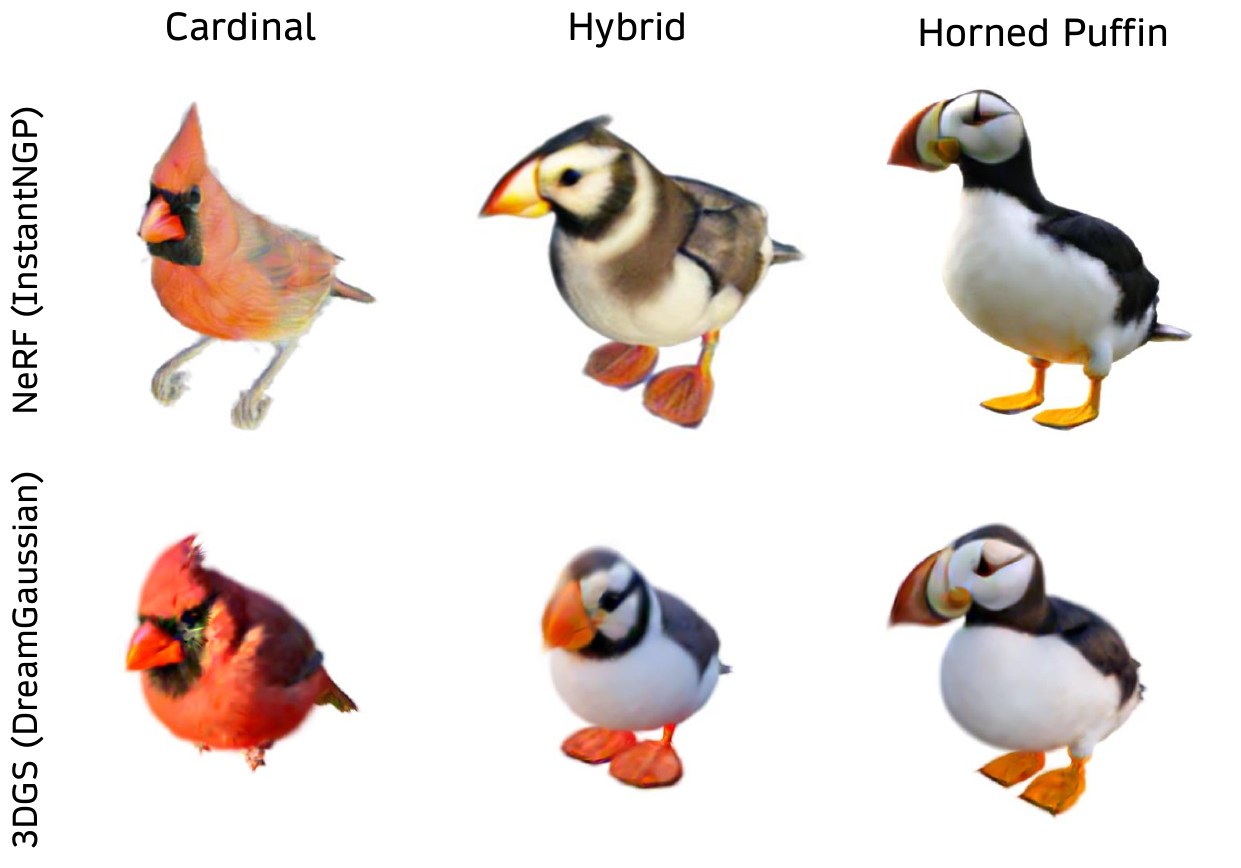}
    \caption{Optimization-based 3D generation with NeRF or 3DGS.}
    \label{fig:nerf3dgs}
\end{figure}

Figure \ref{fig:nerf3dgs} shows that our \ours{} can be used in both NeRF and 3DGS-based 3D generation.

\subsection{InstantMesh}

\begin{figure}[h]
    \centering
    \includegraphics[width=\linewidth]{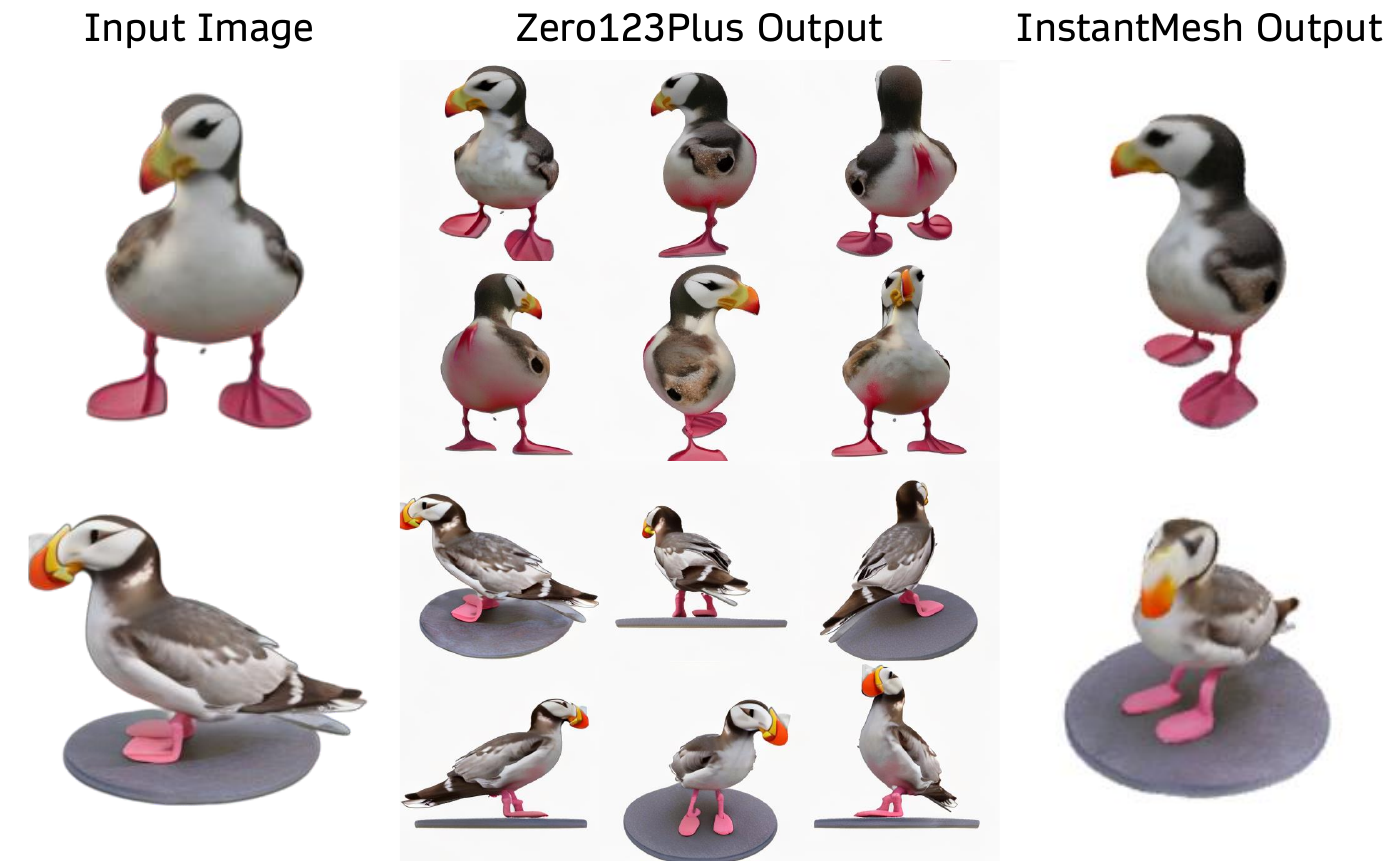}
    \caption{Image-to-3D using front view and side view of generated object.}
    \label{fig:instantmesh}
\end{figure}

We also present additional results in Figure \ref{fig:instantmesh}, where generated images are processed through InstantMesh \cite{xu2024instantmesh} to directly obtain 3D objects. While it provides fast inference, it occasionally produces incorrect conversions as it relies on Zero123Plus \cite{shi2023zero123plus} to predict six views. This limitation arises because Zero123Plus may not have encountered the specific object (or similar objects) during training, leading to inaccurate view predictions. Fine-tuning an image-to-multi-view model typically requires 3D ground-truth data, which is why we focus on text-to-multi-view generation for fine-grained 3D generation. However, with the 3D objects obtained through our method, it may be possible to fine-tune an image-to-multi-view model without relying on 3D ground-truth data—an avenue worth exploring in future research.

\onecolumn
\section{Multi-view generation on common token and fine-grained token} \label{sec:lowgs_analysis}
\begin{figure}[h]
    \centering
    \begin{subfigure}[b]{0.49\textwidth}
        \centering
        \includegraphics[width=0.9\linewidth]{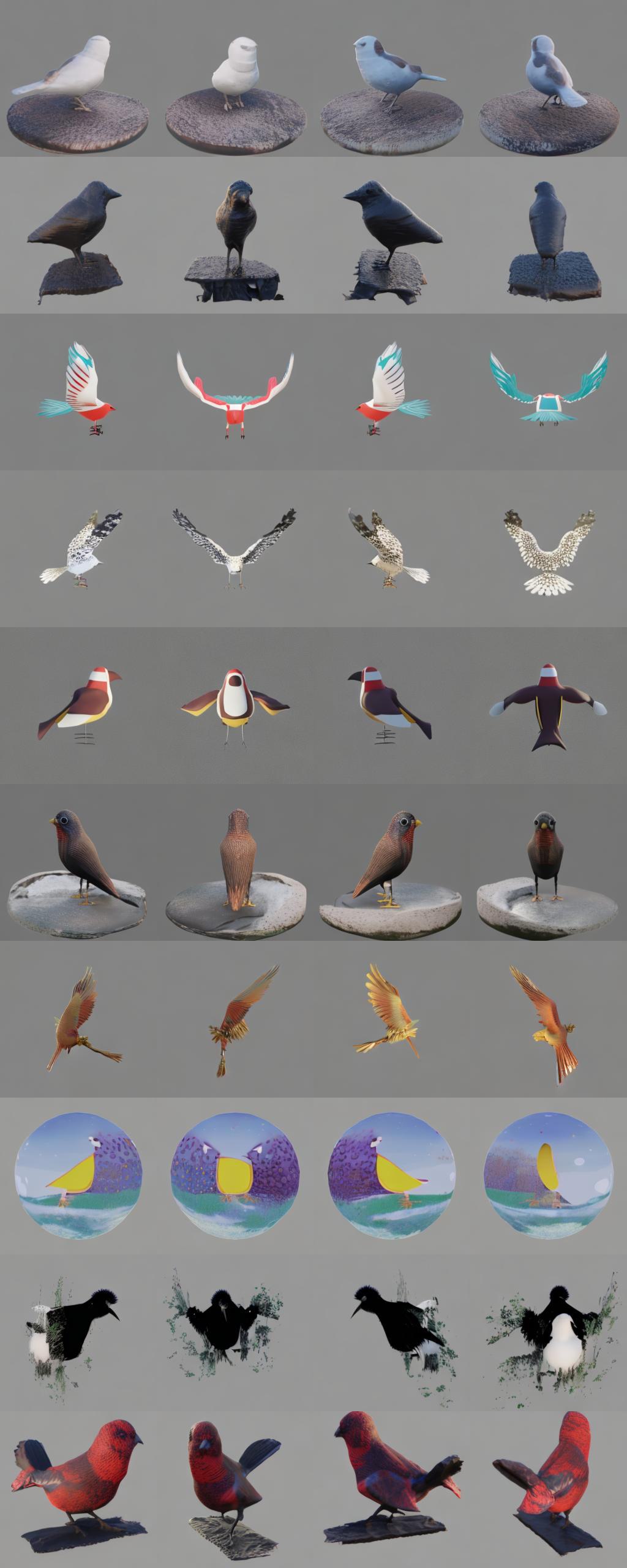}
        \caption{\textit{a bird, 3d asset.}}
    \end{subfigure} \hfill
    \begin{subfigure}[b]{0.49\textwidth}
        \centering
        \includegraphics[width=0.9\linewidth]{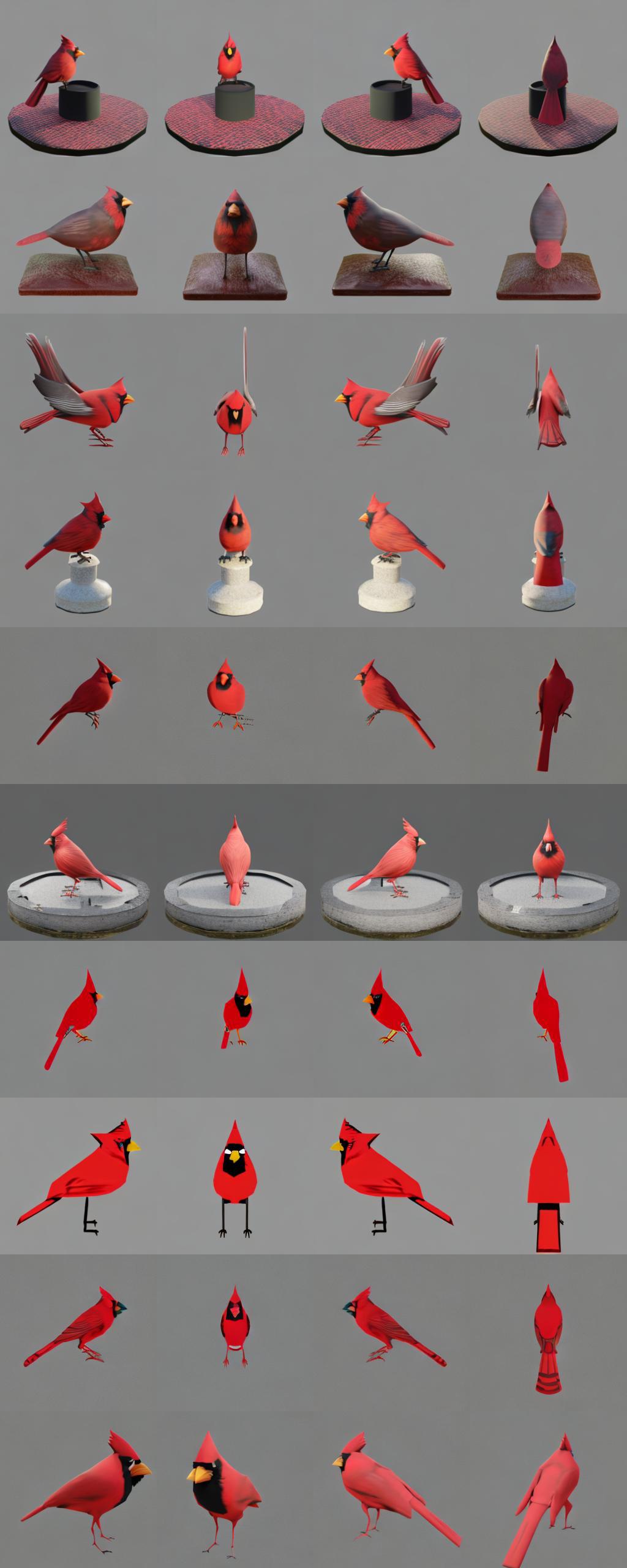}
        \caption{\textit{a cardinal, 3d asset.}}
    \end{subfigure}
    \caption{Multi-view generation with text prompt through MVDream \cite{shi2023mvdream}. The guidance scale is 7.5. Each row is a different seed. \textbf{(a)} The generation varies for different seeds for the token ``bird''. \textbf{(b)} The generation with a fine-grained token ``cardinal''. As highly similar objects are generated for each seed, we can use a lower guidance scale for SDS loss and enable 3D generation without oversaturated effect.}
    \label{fig:lowgs_reason}
\end{figure}

\clearpage

\section{Multi-view generation on existing species} \label{sec:appendix_cls}

\begin{figure}[h]
    \centering
    \begin{subfigure}[b]{0.292\textwidth}
        \centering
        \includegraphics[width=\linewidth]{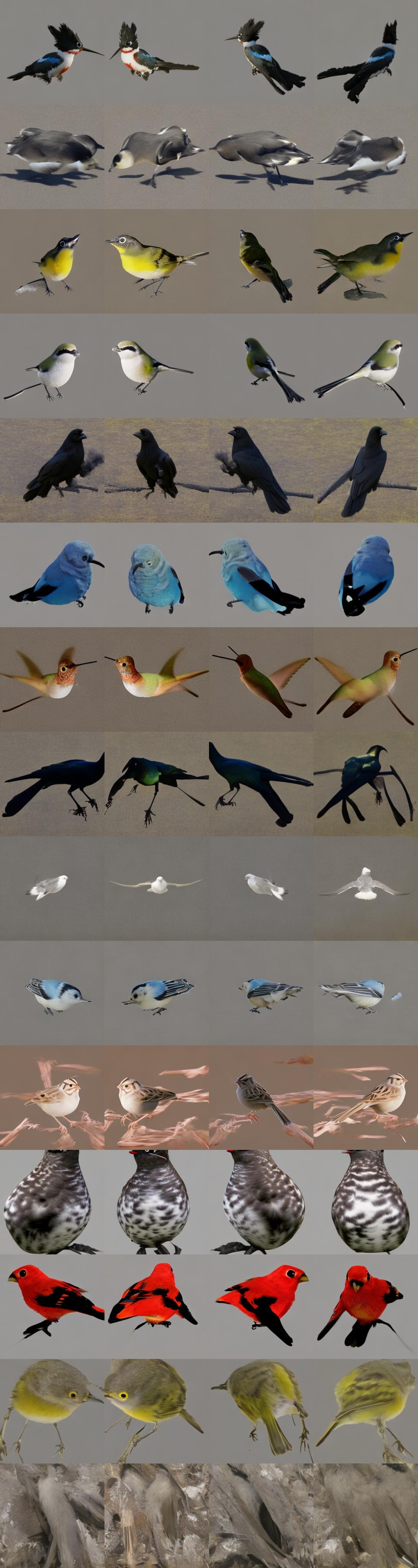}
        \caption{Textual Inversion}
    \end{subfigure} \hfill
    \begin{subfigure}[b]{0.292\textwidth}
        \centering
        \includegraphics[width=\linewidth]{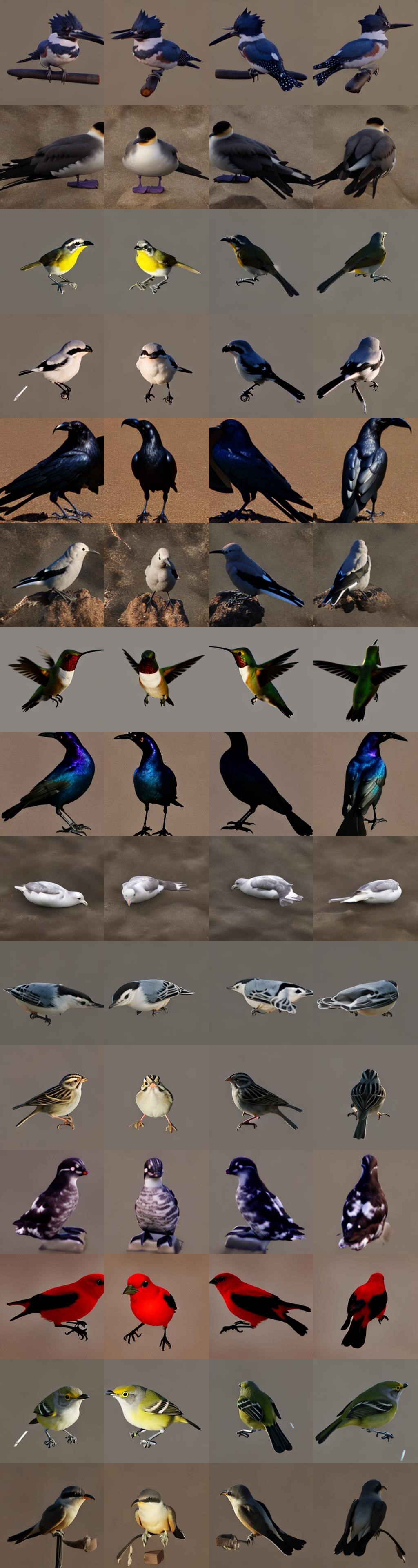}
        \caption{PartCraft}
    \end{subfigure} \hfill
    \begin{subfigure}[b]{0.292\textwidth}
        \centering
        \includegraphics[width=\linewidth]{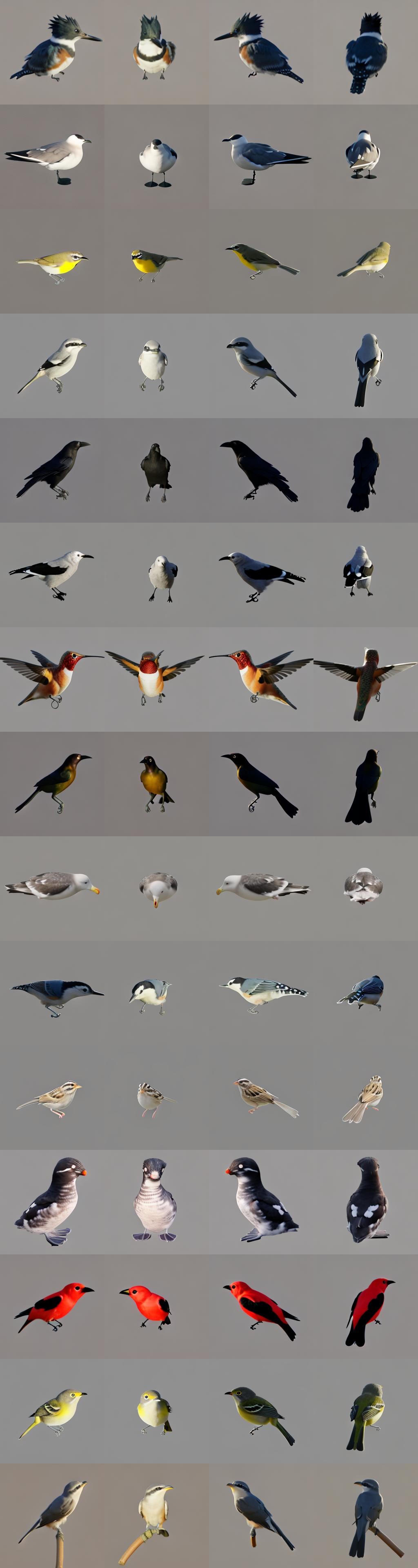}
        \caption{\ours{} (Ours)}
    \end{subfigure} \hfill
    \begin{subfigure}[b]{0.073\textwidth}
        \centering
        \includegraphics[width=\linewidth]{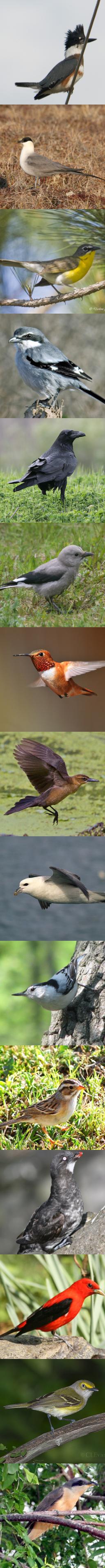}
        \caption{}
    \end{subfigure}
    \caption{Multi-view generation on existing species, trained with respective methods \textbf{(a, b, c)}. \textbf{(d)} One of the training images of the species. Our \ours{} (c) reconstructs well in multi-view perspective compared to Textual Inversion (a) and PartCraft (b), with more consistent orientation and cleaner background.}
    \label{fig:appendix_clscompare}
\end{figure}

\clearpage

\section{Multi-view generation on novel species (random sampling)} \label{sec:appendix_randomsampling}

\begin{figure}[h]
    \centering
    \begin{subfigure}[b]{0.32\textwidth}
        \centering
        \includegraphics[width=0.85\linewidth]{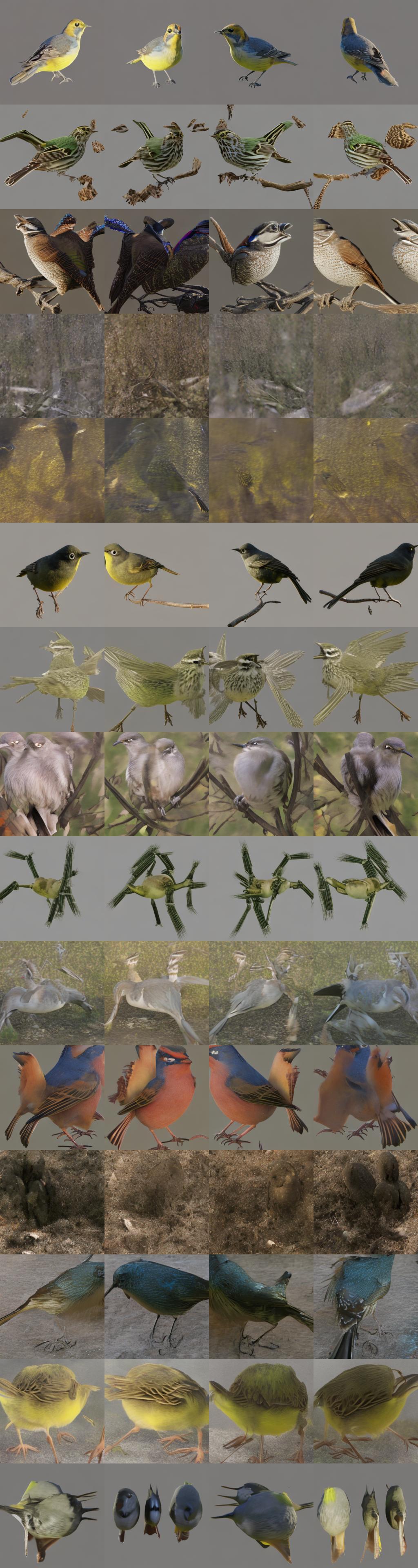}
        \caption{Textual Inversion}
    \end{subfigure} \hfill
    \begin{subfigure}[b]{0.32\textwidth}
        \centering
        \includegraphics[width=0.85\linewidth]{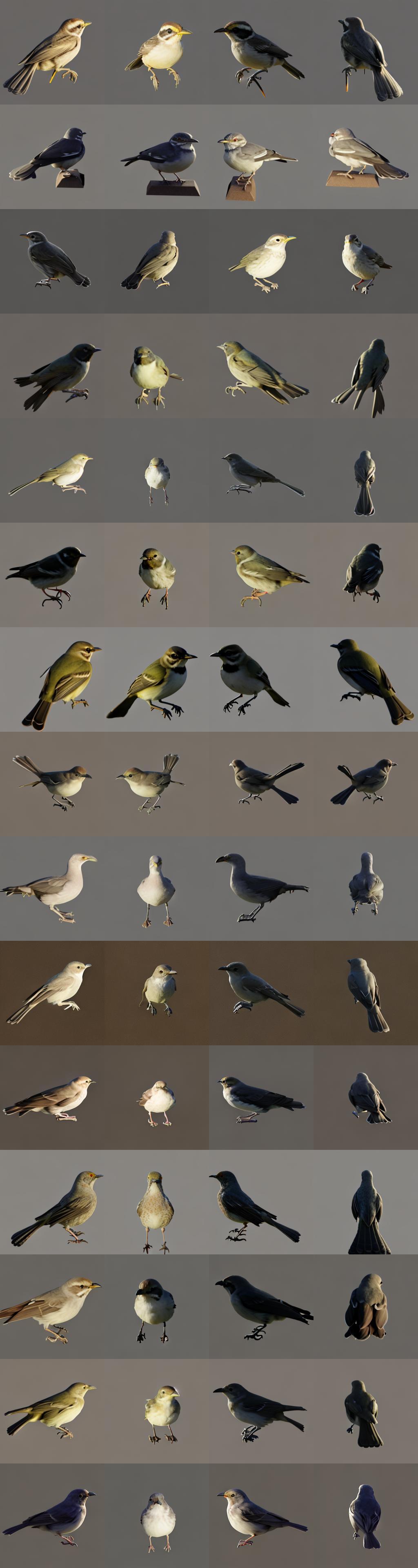}
        \caption{PartCraft}
    \end{subfigure} \hfill
    \begin{subfigure}[b]{0.32\textwidth}
        \centering
        \includegraphics[width=0.85\linewidth]{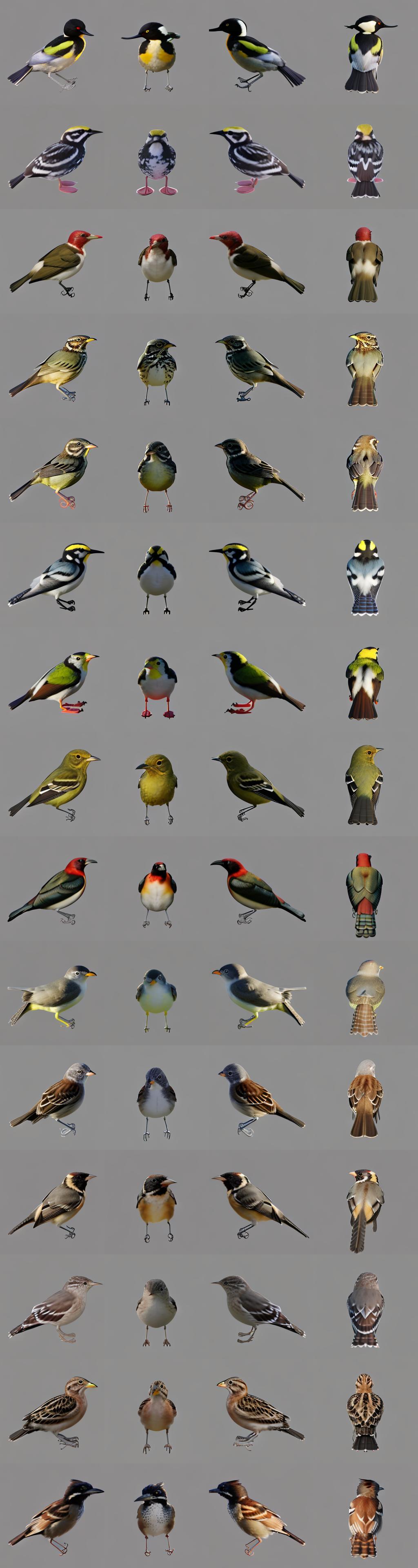}
        \caption{\ours{} (Ours)}
    \end{subfigure}
    \caption{Multi-view generation on novel species (random sampling), trained with respective methods. All were generated with the same seed but with different sampled part latents. \textbf{(a)} Trained with Textual Inversion, the generated images are often incomprehensible, indicating that direct sampling from word embedding space is insufficient to generate novel species. \textbf{(b)} PartCraft has a non-linear projector to project word embeddings, while able to generate comprehensible objects, but lacking diversity since it is not trained to have a continuous distribution of part latents. \textbf{(c)} Our \ours{} not only can generate images of diverse species, also stable in terms of bird pose.}
    \label{fig:appendix_randomcompare}
\end{figure}

\clearpage

\section{Multi-view generation on novel species (interpolation)} \label{sec:appendix_interpolation}

\begin{figure}[h]
    \centering
    \begin{subfigure}[b]{\textwidth}
        \centering
        \includegraphics[width=\linewidth]{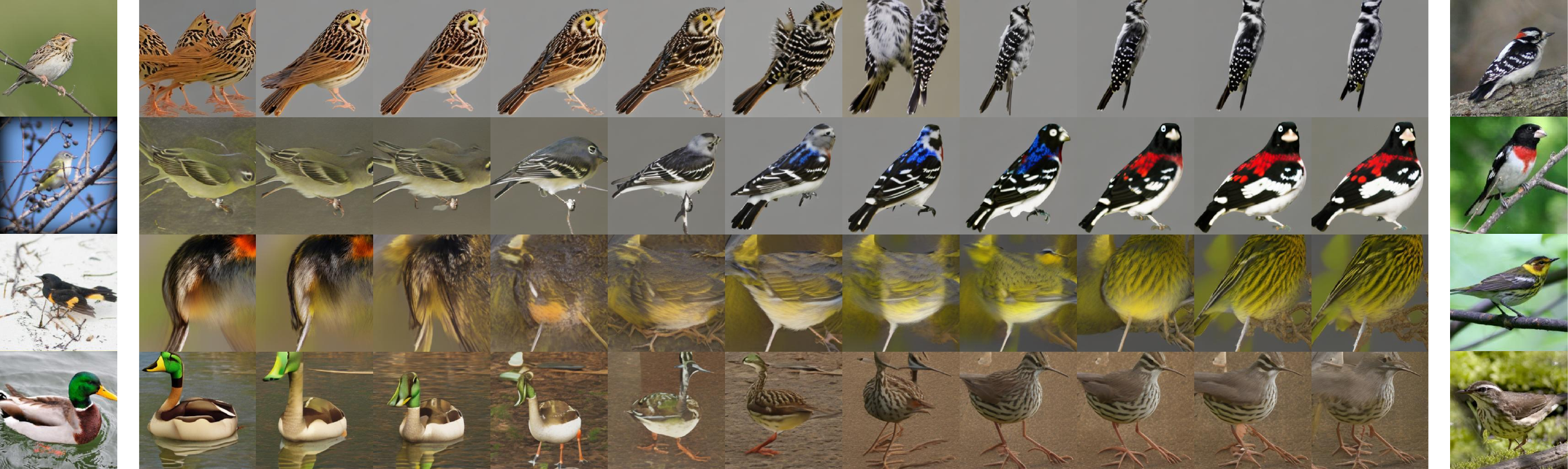}
        \caption{Textual Inversion}
    \end{subfigure}
    \begin{subfigure}[b]{\textwidth}
        \centering
        \includegraphics[width=\linewidth]{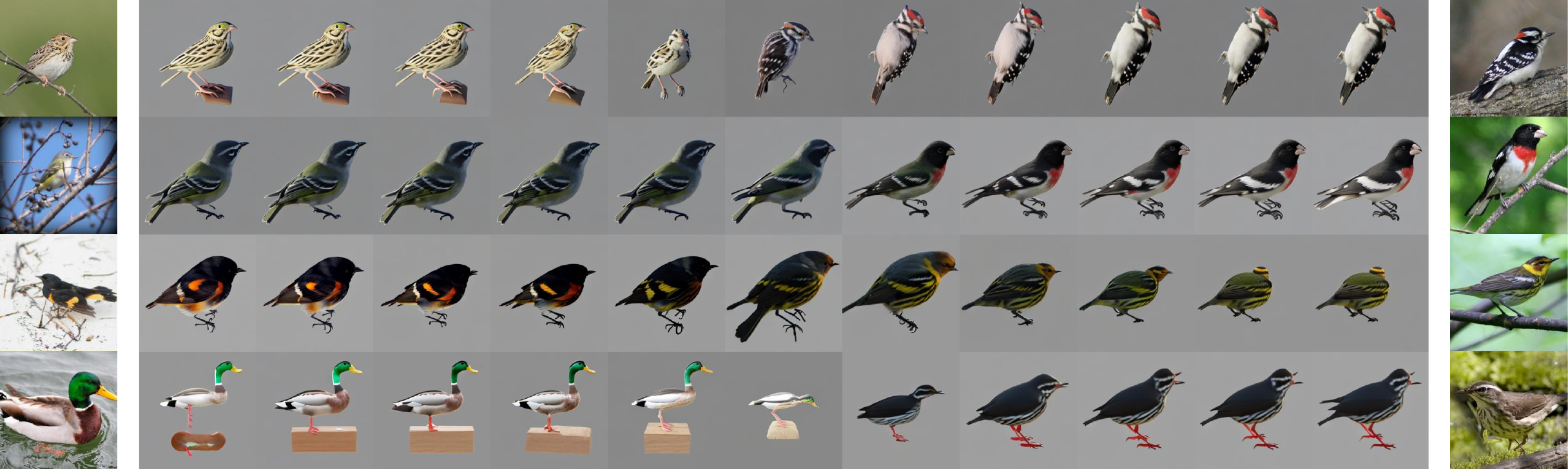}
        \caption{PartCraft}
    \end{subfigure}
    \begin{subfigure}[b]{\textwidth}
        \centering
        \includegraphics[width=\linewidth]{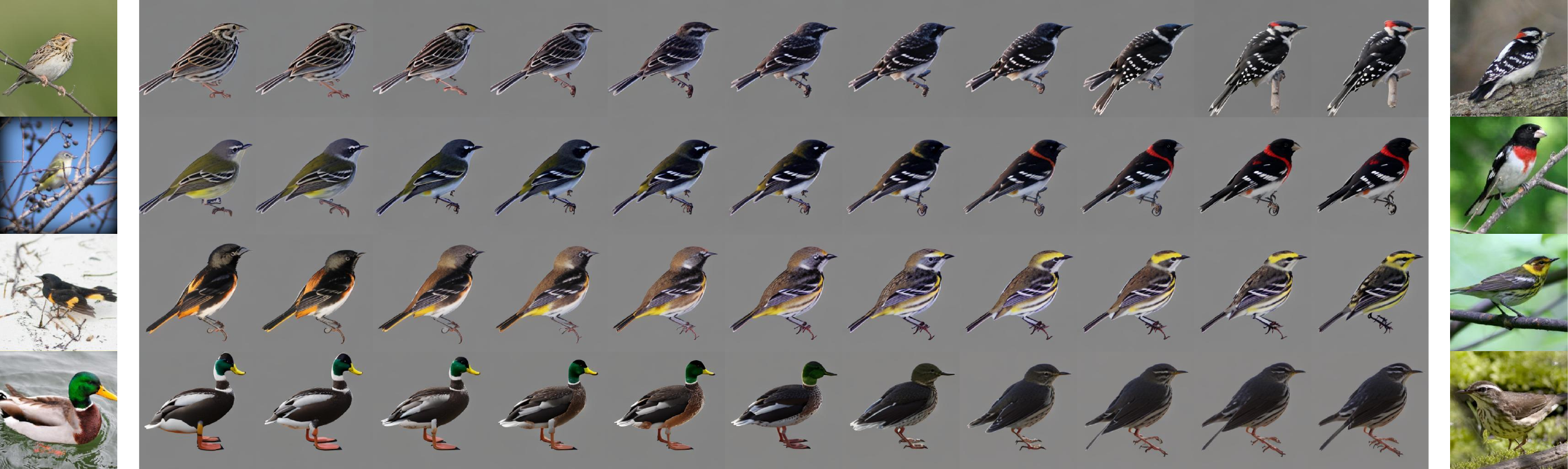}
        \caption{\ours{} (Ours)}
    \end{subfigure}
    \caption{Multi-view generation on novel species (interpolation) trained with respective methods. All images were generated using the same seed, but with different interpolated part latents. \textbf{(a)} Trained with Textual Inversion, the generated images are often incomprehensible, consistent with previous visualizations. \textbf{(b)} PartCraft exhibits an abrupt switching effect, with the object remaining unchanged before and after switching, as it is not designed to support a continuous distribution of part latents. \textbf{(c)} Our \ours{} method successfully generates interpolated samples with gradual visual changes.}
    \label{fig:appendix_interpolationsample}
\end{figure}

\clearpage

\section{Qualitative comparisons of visual coherency before and after applying feature consistency loss} \label{sec:appendix_beforeafter_cl}

\begin{figure}[h]
    \centering
    \begin{subfigure}[b]{0.49\textwidth}
        \centering
        \includegraphics[width=0.9\linewidth]{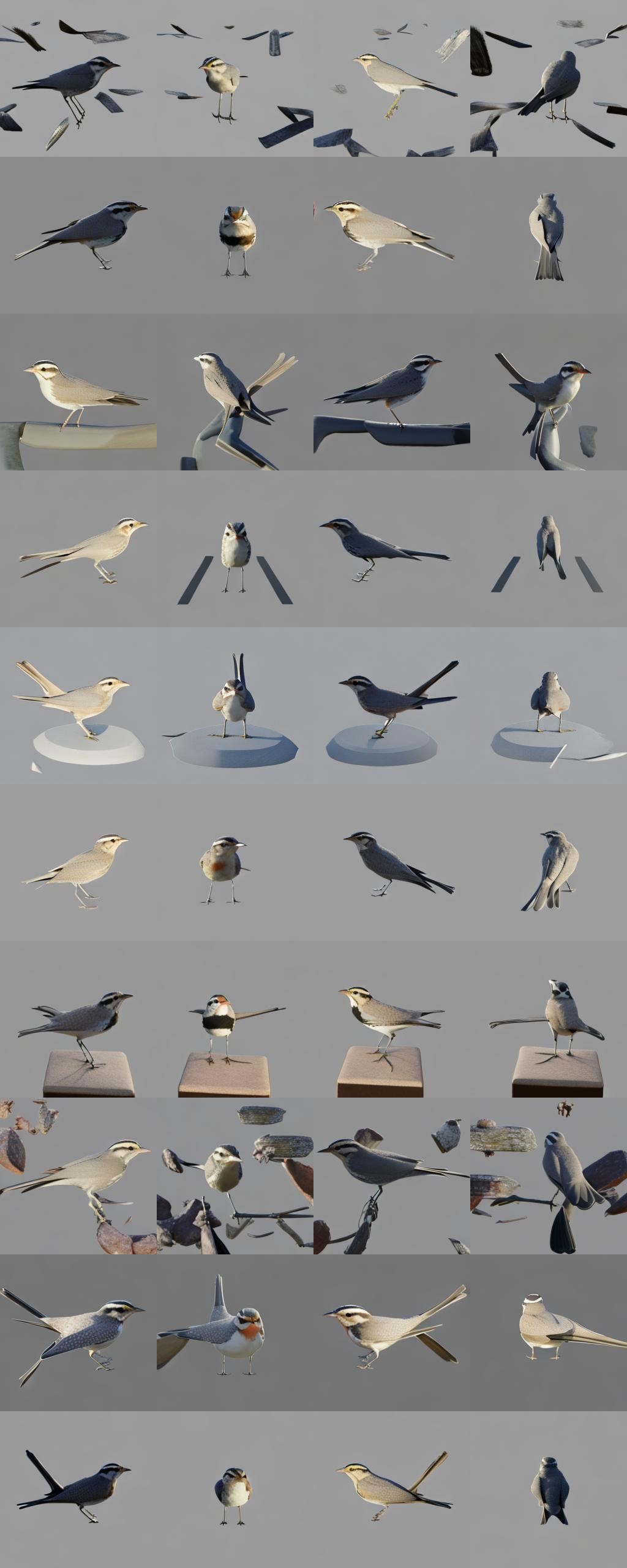}
        \caption{}
    \end{subfigure} \hfill
    \begin{subfigure}[b]{0.49\textwidth}
        \centering
        \includegraphics[width=0.9\linewidth]{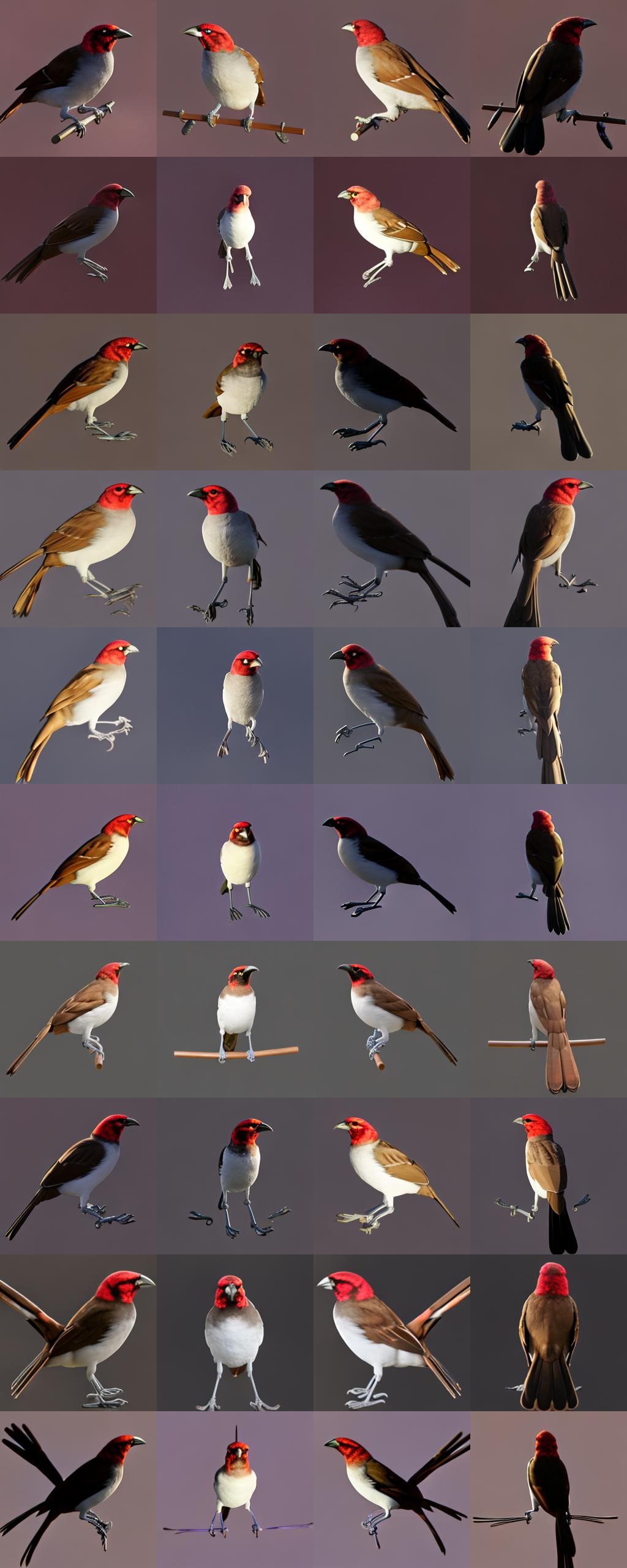}
        \caption{}
    \end{subfigure}
    \caption{Each row is a different seed with a random sampled part latents \textbf{(a)} before applying $\mathcal{L}_{\text{cl}}$ and \textbf{(b)} after applying. We can see that, although the part latents are unseen during training, applying the loss can increase visual coherency and reduce artifacts.}
    \label{fig:appendix_after_cl}
\end{figure}

%% file: tab_fig/ab.tex
\begin{table}[t]
    \centering
    \adjustbox{max width=\linewidth}{
        \begin{tabular}{l|ccccccccc}
        Method &$\mathcal{L}_\text{diff}$&$\mathcal{L}_\text{attn}$&$\mathcal{L}_\text{cl}$&$\mathcal{L}_\text{reg}$&$\mathbf{s}$&f&$\mathbf{p}$&g&$\mathbf{t}$\\
        \toprule
        \TI{}~\cite{gal2022textualinversion}  & \cmark &\cmark &\xmark&\xmark&\xmark&\xmark&\xmark&\xmark &\cmark\\
        \PC{}~\cite{ng2024partcraft} & \cmark &\cmark &\xmark&\xmark&\xmark&\xmark&\xmark&\cmark &\cmark\\
        \bf \ours{} &\cmark &\cmark &\cmark&\cmark&\cmark&\cmark&\cmark&\cmark &\xmark\\
        \end{tabular}
    }
    \vspace{-2mm}
    \caption{Comparison among \TI{}, \PC{} and \ours{}. $s$ denotes species embeddings. $f$ denotes mapping from species embedding to latent space. $p$ is part latent space. $g$ is projector from latent space to textual embedding. $t$ is learnable textual embeddings. For \ours{}, $t$ is from $g$ with input $p$. For \PC{}, $t$ is from $g$ with input $t$. For \TI{}, it optimizes $t$.}
    \vspace{-0.3cm}
    \label{tab:comparison}
\end{table}